\documentclass[10pt,twocolumn,letterpaper]{article}

\usepackage[pagenumbers]{cvpr} % To force page numbers, e.g. for an arXiv version

\usepackage{graphicx}
\usepackage{amsmath}
\usepackage{amssymb}
\usepackage{amsfonts}
\usepackage{booktabs}

\DeclareMathOperator*{\argmin}{arg\,min}

\usepackage{cuted}
\usepackage{capt-of}
\usepackage{dsfont}

\usepackage{rotating}

\usepackage[pagebackref,breaklinks,colorlinks]{hyperref}

\usepackage[capitalize]{cleveref}
\crefname{section}{Sec.}{Secs.}
\Crefname{section}{Section}{Sections}
\Crefname{table}{Table}{Tables}
\crefname{table}{Tab.}{Tabs.}

\def\confYear{2022}

\begin{document}

\title{Discovering Multiple and Diverse  Directions for Cognitive Image Properties}

\renewcommand{\thefootnote}{\fnsymbol{footnote}}

\author{\stepcounter{footnote}\vspace{1mm}  Umut Kocasari$^{1}$
\hspace{0.75cm}
Alperen Bag$^{2}$ 
\hspace{1cm}
Oğuz Kaan Yüksel$^{3}$
\hspace{1cm}
Pinar Yanardag$^{1}$
\\
$^1$Boğaziçi University
\hspace{2em} 
$^2$TUM
\hspace{2em} 
$^3$EPFL
\\
{\tt\small umut.kocasari@boun.edu.tr}
\hspace{0.75em}
{\tt\small alperen.bag@tum.de}
\hspace{0.75em}
{\tt\small oguz.yuksel@epfl.ch}
\hspace{0.75em}
{\tt\small yanardag.pinar@gmail.com}
}

\maketitle

\vspace*{-\baselineskip}
\begin{strip} 
%\vspace{-1.2cm}
\qquad \qquad  \, \,  $\alpha$- -  \qquad \, \, $\alpha$=0 \quad  \qquad $\alpha$++ \qquad \, \quad   $\alpha$- - \, \, \qquad $\alpha$=0 \; \, \qquad $\alpha$++ \qquad  \quad  $\alpha$- - \; \; \, \qquad $\alpha$=0 \, \, \, \qquad $\alpha$++
\vspace{-0.15cm} 
  \begin{center}
 \begin{turn}{90} \quad Ours (1) \; \, \, Ours (2) \; \; Ours (3) \; \, GANalyze \end{turn}
        \includegraphics[width=1.9\columnwidth]{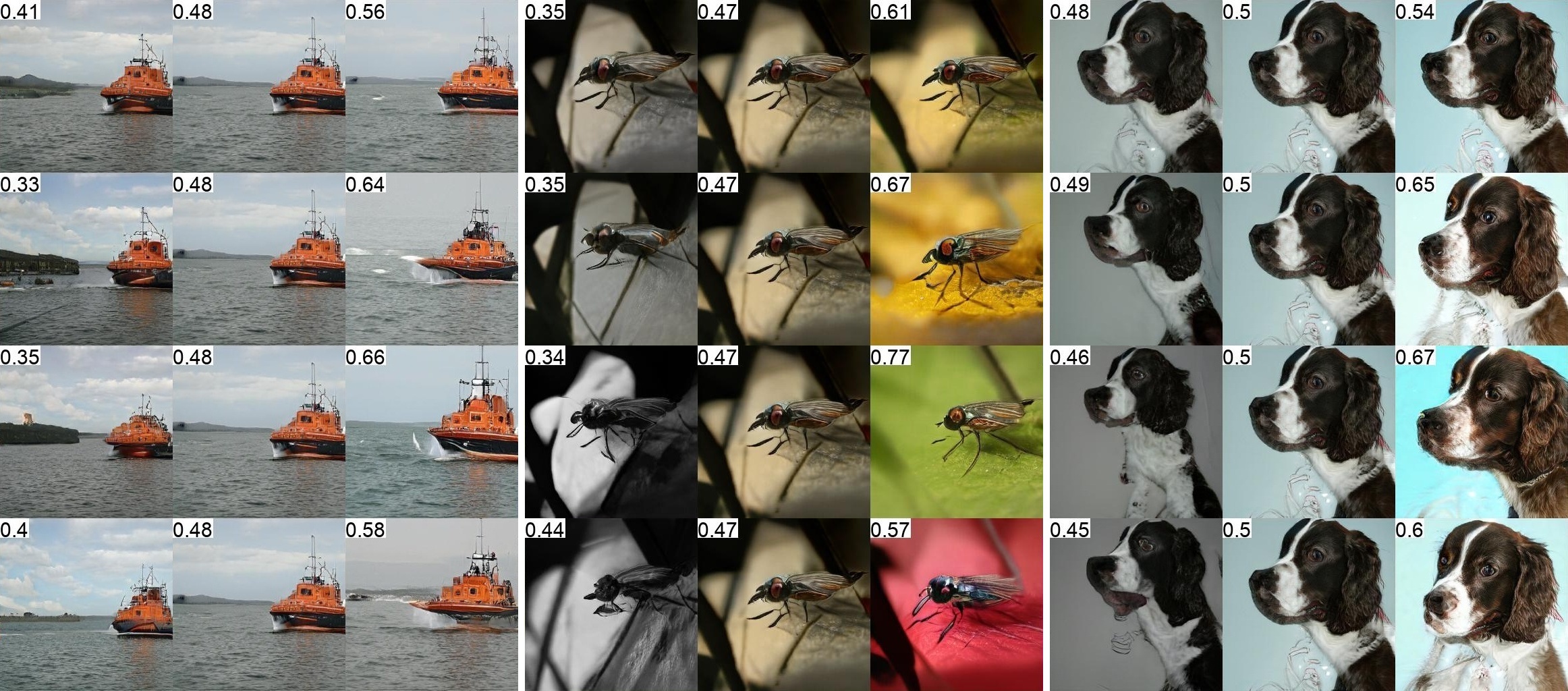}
        
        \hspace{0.3cm} Memorability \hspace{3.0cm} Emotional Valence \hspace{3.0cm} Aesthetics
       % \vspace{-0.75cm} 
 \captionof{figure}{Examples of image manipulations for Memorability, Emotional Valence, and Aesthetics produced by our method (labeled \textit{Ours}) and GANalyze \cite{goetschalckx2019ganalyze}. Our method finds multiple directions that produce diverse outputs while achieving similar effect on model scores. $\alpha=0$ denotes the input image, while $\alpha$- - and $\alpha$++ shift the latent code towards negative and positive directions, respectively. The classifier scores for each image are displayed in the upper left corner.}
\label{fig:teaser}
\end{center}
\end{strip}
 
\begin{abstract}

Recent research has shown that it is possible to find interpretable directions in the latent spaces of pre-trained GANs. These directions enable controllable generation and support a variety of semantic editing operations. While previous work has focused on discovering a single direction that performs a desired editing operation such as \textit{zoom-in}, limited work has been done on the discovery of \textit{multiple} and \textit{diverse} directions that can achieve the desired edit. In this work, we propose a novel framework that discovers multiple and diverse directions for a given property of interest. In particular, we focus on the manipulation of cognitive properties  such as \textit{Memorability, Emotional Valence} and \textit{Aesthetics}. We show with extensive experiments that our method successfully manipulates these properties while producing diverse outputs.  Our project page and source code can be found at \url{http://catlab-team.github.io/latentcognitive}.
\end{abstract}
 
\vspace{-0.25cm} 
\section{Introduction}
\label{sec:introduction}
Generative Adversarial Networks (GANs) \cite{NIPS2014_5423} are popular methods due to their success in generating photorealistic images, and have become a widely used technique for high-quality results thanks to their ability to generate high-resolution images. Image manipulation \cite{wang2017highresolution}, image generation \cite{DBLP:journals/corr/abs-1710-10916}, image super-resolution \cite{Sun_2020}, image-to-image translation \cite{CycleGAN} are some of their creative use cases. Although it is possible to generate realistic images, GAN models often lack fine-grained control over the generated images. Some degree of control can be achieved by training them under certain conditions \cite{mirza2014conditional}, but this approach is limited because it requires labeled data. One of the earlier approaches such as \cite{radford2015unsupervised} attempts to control the generated images by interpolating latent vectors. However, this approach is quite limited in terms of the possible manipulations that can be achieved. Recently, several methods have been proposed to provide fine-grained control over latent space \cite{jahanian2019steerability,voynov2020unsupervised,shen2020interfacegan,harkonen2020ganspace}. Most of these works find domain-independent and general directions such as \textit{rotation} or \textit{zoom-in} \cite{plumerault2020controlling,jahanian2019steerability}, while others proposed techniques to explore domain-specific directions, such as changing \textit{age, gender} or \textit{expression} on facial images \cite{shen2020interfacegan}. In general, the learned directions are used to change the properties of images by shifting the latent code to a desired extent in the learned directions. After modifying the latent code, the particular property of the image is changed in that direction.

However, most previous work on the manipulation of latent space has focused on the discovery of a \textit{single} direction by which the desired editing can be achieved, such as  \textit{zoom-in} \cite{jahanian2019steerability,harkonen2020ganspace} or altering facial features, such as applying a \textit{lipstick} \cite{Kocasari2021StyleMCMB,yuksel2021latentclr}. In this work, we focus on the discovery of \textit{multiple} and \textit{diverse} directions that can achieve the desired edit. In particular, we focus on discovering multiple directions for cognitive properties such as \textit{memorability, emotional valence, and aesthetics}, where achieving a desired edit in multiple different ways can be beneficial. For cognitive features, there is no precise definition of what they entail \cite{goetschalckx2019ganalyze}. For example, while it is possible to write editing functions that perform  \textit{zoom-in}  or \textit{transformation} operations on the image \cite{jahanian2019steerability}, we do not have a concrete definition for what \textit{memorability} is. GANalyze \cite{goetschalckx2019ganalyze} proposes a framework to discover latent directions for high-level cognitive attributes, which navigates the latent space to manipulate images by taking a step in the direction that increases or decreases a cognitive property. Their method finds a single direction using an optimization framework that benefits from off-the-shelf classifiers. However, instead of fixing a single direction for a cognitive property of interest, we show that there can be more than one way to manipulate an input image toward a cognitive property, e.g., modifying it to be perceived more \textit{emotional} (see Figure \ref{fig:teaser} for multiple directions found by our method).

In this work, we propose a novel framework for discovering multiple and diverse directions for cognitive properties. We apply our method to cognitive properties such as \textit{emotional valence, memorability} and \textit{aesthetics}. Our quantitative and qualitative experiments show that our method outperforms GANalyze \cite{goetschalckx2019ganalyze} and provides better manipulations on cognitive features.  As far as we know, our framework is the first to propose  multiple directions for discovering cognitive properties.

The rest of this paper is organized as follows. Section \ref{sec:related_work} discusses related work on latent space manipulation. Section \ref{sec:methodology} presents our framework and Section \ref{sec:experiments} discusses quantitative and qualitative results. Section \ref{sec:limitations} discusses the limitations of our work and Section \ref{sec:conclusion} concludes the paper.
\section{Related Work}
\label{sec:related_work}
In this section, we discuss related work on generative adversarial networks, unsupervised and supervised methods for manipulating latent space.

\subsection {Generative Adversarial Networks} 
Generative Adversarial Networks (GANs) are two-part networks that model the real world into generative space using Deep Learning methods \cite{NIPS2014_5423}. The generative part of the network tries to generate images that are similar to the dataset, while the adversarial part tries to detect whether the generated image is from the training dataset or  generated. The main goal of GANs is to model the image space such that the generated images are indistinguishable from the images in the dataset. Generative networks can generate realistic images from random noise vectors, called latent vectors, through internal mapping. Recently, large-scale GAN models such as BigGAN \cite{BigGAN} and StyleGAN \cite{StyleGAN} have paved the way for generating photorealistic images. BigGAN \cite{BigGAN} is a comprehensive GAN model trained on ImageNet \cite{russakovsky2015imagenet} that supports image generation in multiple categories due to its conditional architecture. StyleGAN \cite{StyleGAN} is another popular GAN approach where the generator attempts to match the input latent code to an intermediate latent space using a mapping network.

\subsection{Latent Space Manipulation}
Recently, it has been shown that the latent space of GANs contains interpretable directions \cite{StyleGAN}  to manipulate images in various ways, such as performing image transformations like \textit{zoom-in, shift} or changing attribute-based edits like changing \textit{hair color, age} or \textit{gender} \cite{StyleGAN}. Several techniques have been proposed to exploit the latent structure of GANs, and they are mainly divided into two areas. 

\paragraph{Supervised methods} Supervised latent space manipulation approaches typically benefit from pre-trained classifiers to guide the optimization process and discover meaningful directions, or use labeled data to train new classifiers that directly target  directions of interest. \cite{goetschalckx2019ganalyze} uses an externally trained classifier called \textit{assessor} functions to discover directions for cognitive image features in the latent space of BigGAN. Its optimization process is guided by getting feedback from the assessor, and the resulting optimal direction can be used to control the properties of an image. \cite{shen2020interfacegan} is another supervised method that uses labeled data such as \textit{age, gender} and \textit{facial expression}. They train a Support Vector Machine (SVM) \cite{noble2006support} classifier and use the normal vector of the obtained hyperplane as the latent direction. Similarly, \cite{abdal2021styleflow} uses conditional continuous normalization flows to perform supervised attribute processing in the latent space of StyleGAN2. Recently, text-based manipulation methods have been proposed \cite{patashnik2021styleclip,Kocasari2021StyleMCMB}, which use CLIP \cite{Radford2021LearningTV} to perform fine-grained and disentangled manipulations of images.

\paragraph{Unsupervised methods} Exploring meaningful directions in latent space can also be done in an unsupervised manner. GANspace \cite{harkonen2020ganspace} applies principal component analysis (PCA) \cite{wold1987principal} to randomly sampled latent vectors of intermediate layers of BigGAN and StyleGAN models. Their framework uses the resulting principal components as latent directions and discovers several meaningful manipulations such as changing \textit{gender, hair color or age} on facial images. A similar approach is used by \cite{shen2020closed}, where they directly optimize the intermediate weight matrix of the GAN model in closed form. They compute the eigenvectors of the matrix of the first projection step and selecting the eigenvectors with the largest eigenvalues, they obtain semantically meaningful directions in the latent space. \cite{voynov2020unsupervised} proposes an unsupervised approach to discover interpretable directions. It uses a classifier-based approach that attempts to identify a set of directions corresponding to different image transformations. A more recent work, LatentCLR \cite{yuksel2021latentclr}, proposes a contrastive learning approach to find unsupervised directions that are transferable to different classes.

\paragraph{Conditional Directions} The manipulation directions can also be found conditionally using a neural network. \cite{zhuang2021enjoy} uses local transformation to find a direction conditioned by the latent vector. \cite{tzelepis2021warpedganspace} uses a warping network to learn nonlinear paths in the latent space. \cite{khrulkov2021latent} proposes a framework that works for domains with more complex non-textured factors of variation using a trainable Neural ODE to find nonlinear directions. \cite{jahanian2019steerability} proposes a self-supervised method that benefits from task-specific editing functions. They first apply an editing operation to the original image, such as a \textit{rotation}. Then they learn a direction for that specific operation by minimizing the distance between the original image and the edited image.  Their method uses a neural network based approach to perform a nonlinear walk with fixed step sizes in the latent space instead of a variable step size, a requirement that arises from the need to achieve transformations of different sizes based on the feature of interest, e.g. \textit{rotation}.

The closest work in terms of multiple direction discovery is \cite{gu2020image}, where a GAN inversion method called mGANprior is proposed to reconstruct a given image. Their method generates multiple feature maps on a given intermediate layer of the generator by using multiple latent codes and fusing them with adaptive channel meaning to recover the input image. However, their goal is GAN inversion and their methodology differs significantly from ours.

\begin{figure*}[!htb]
\begin{center}
    \begin{tabular}{ l }
        \rule{0pt}{50pt}\rotatebox{90}{\quad \quad \quad \; Single Direction } 
    
  \includegraphics[width=0.32\linewidth]{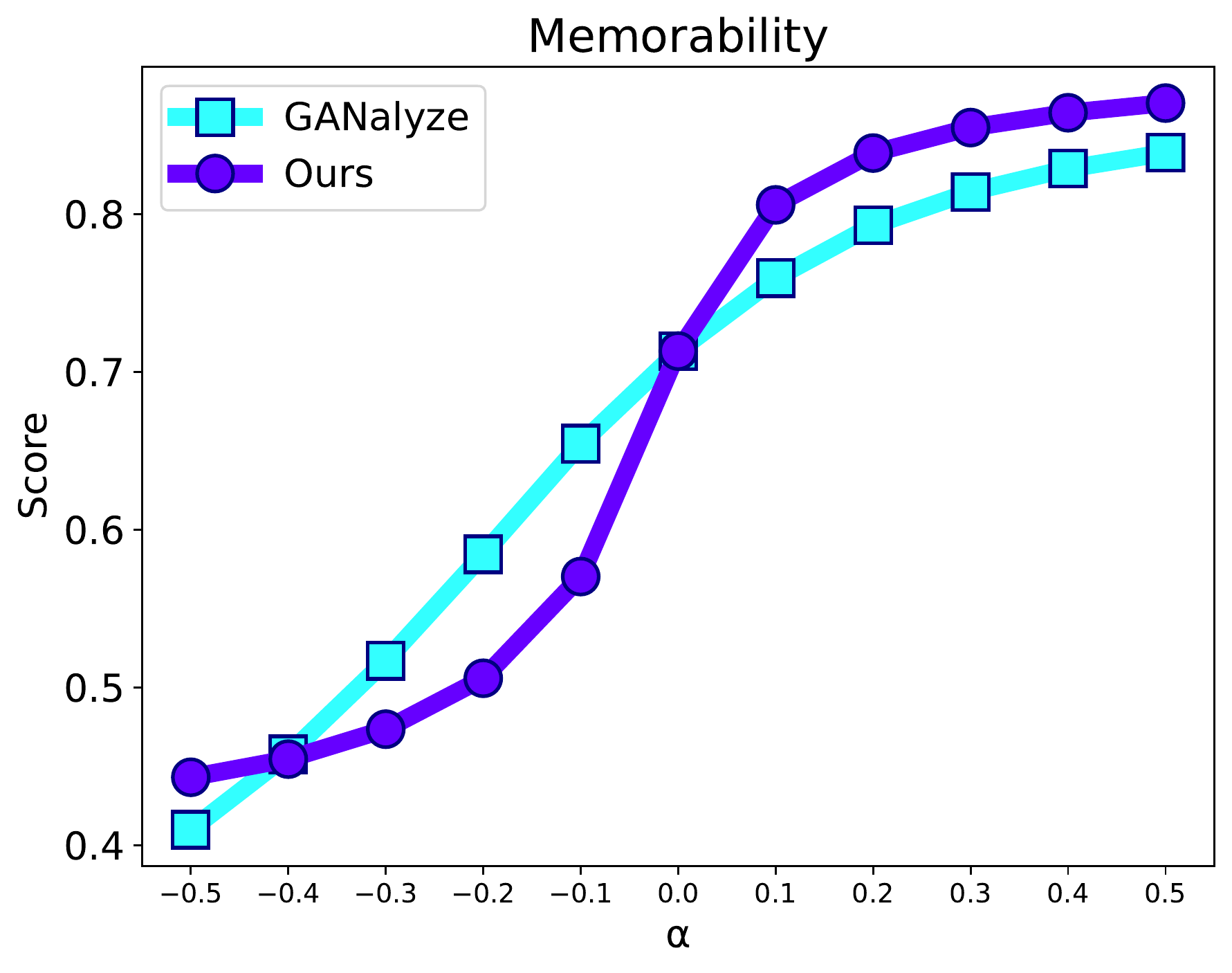}
  \includegraphics[width=0.32\linewidth]{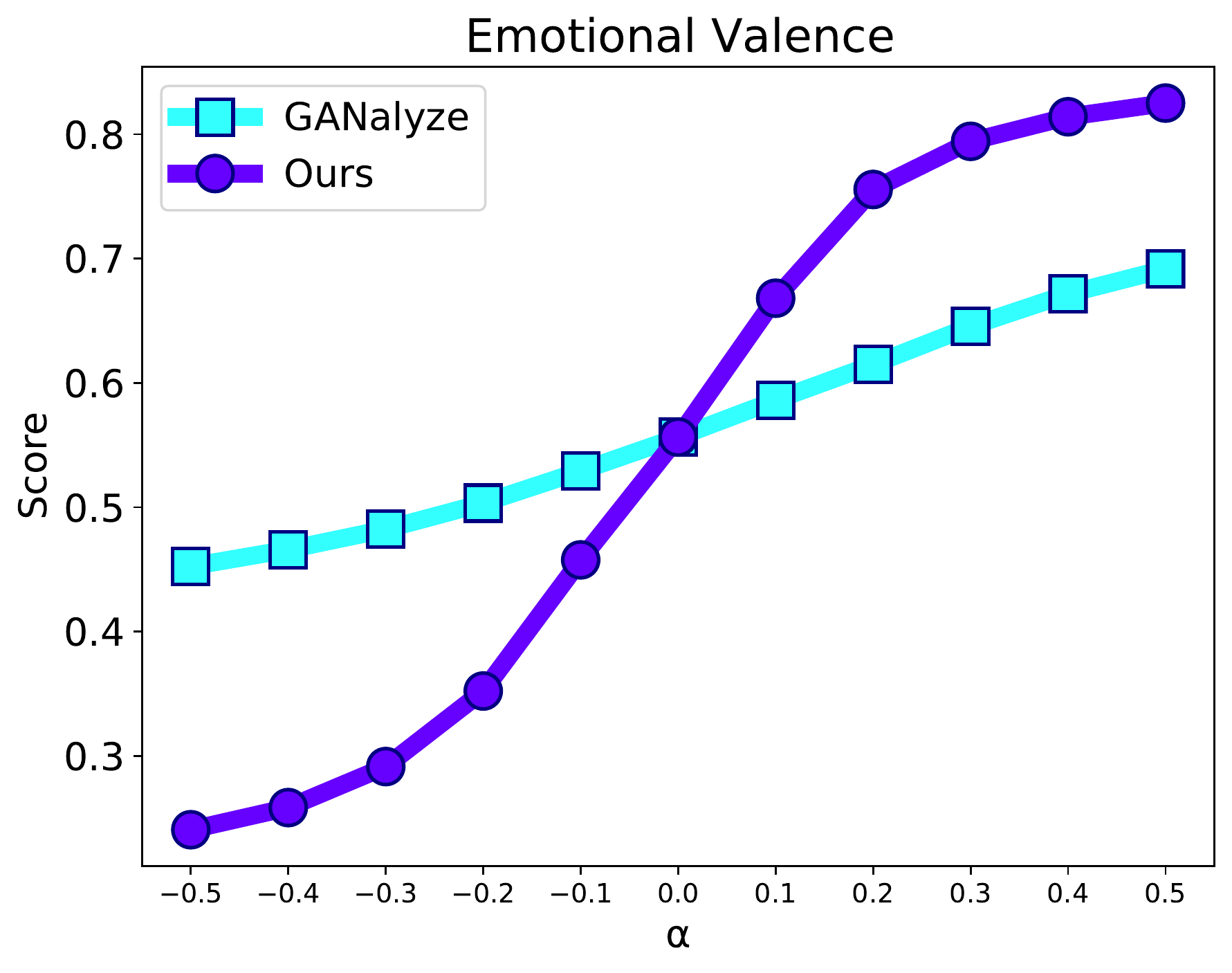}
  \includegraphics[width=0.32\linewidth]{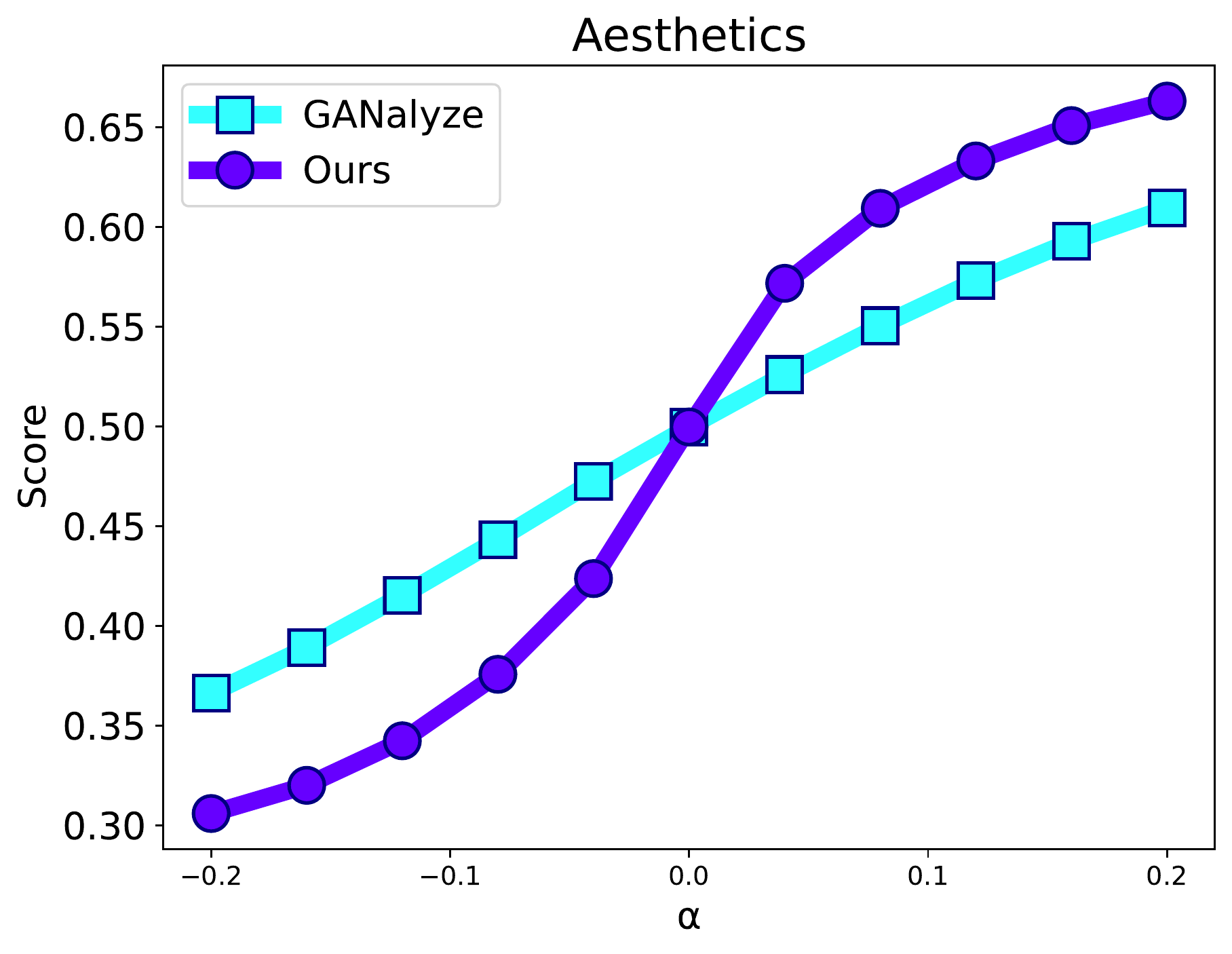}
        \end{tabular}
  \begin{tabular}{ l }
        \rule{0pt}{50pt}\rotatebox{90}{\qquad \, \, Multiple Directions } 
    
  \includegraphics[width=0.32\linewidth]{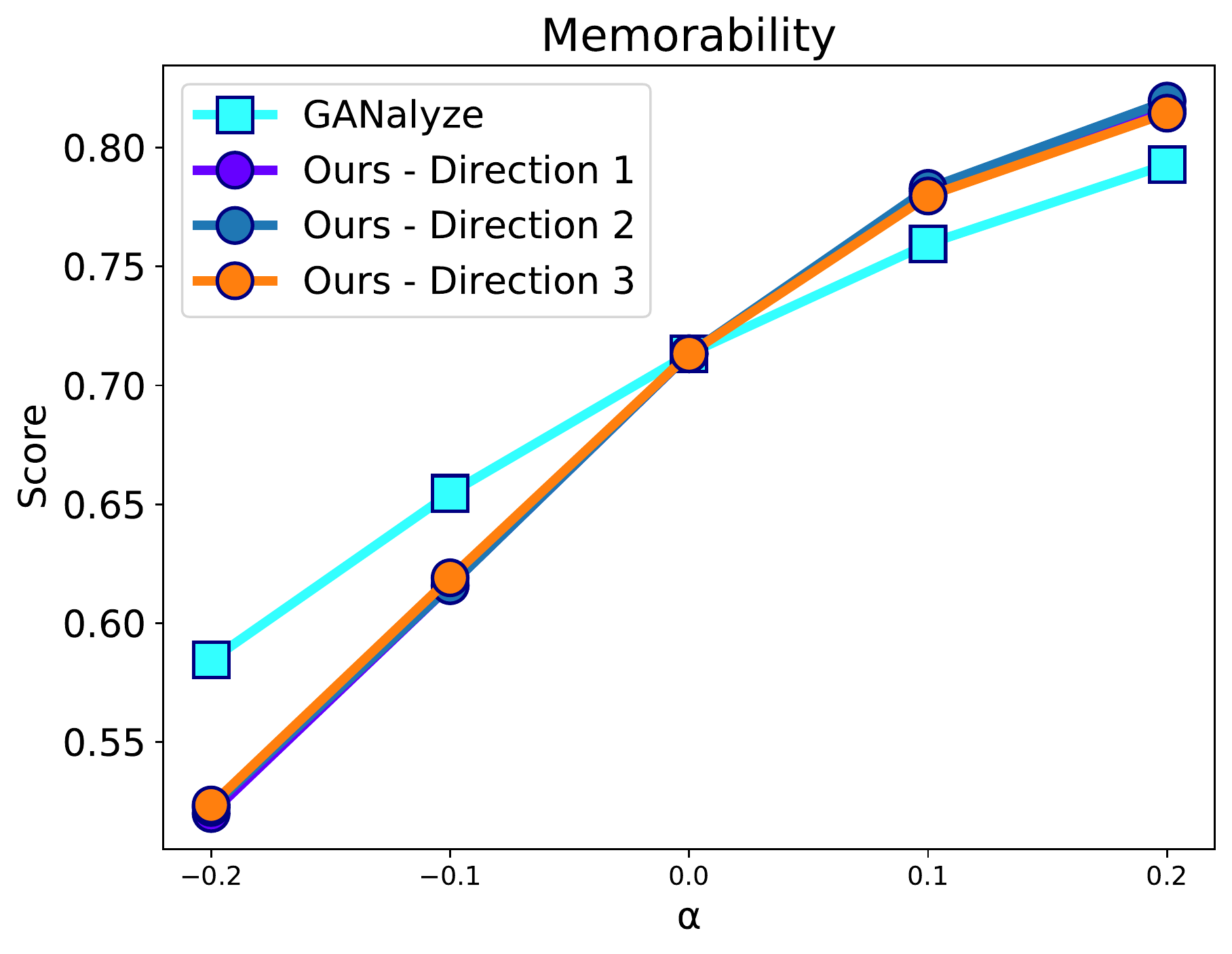}
 \includegraphics[width=0.32\linewidth]{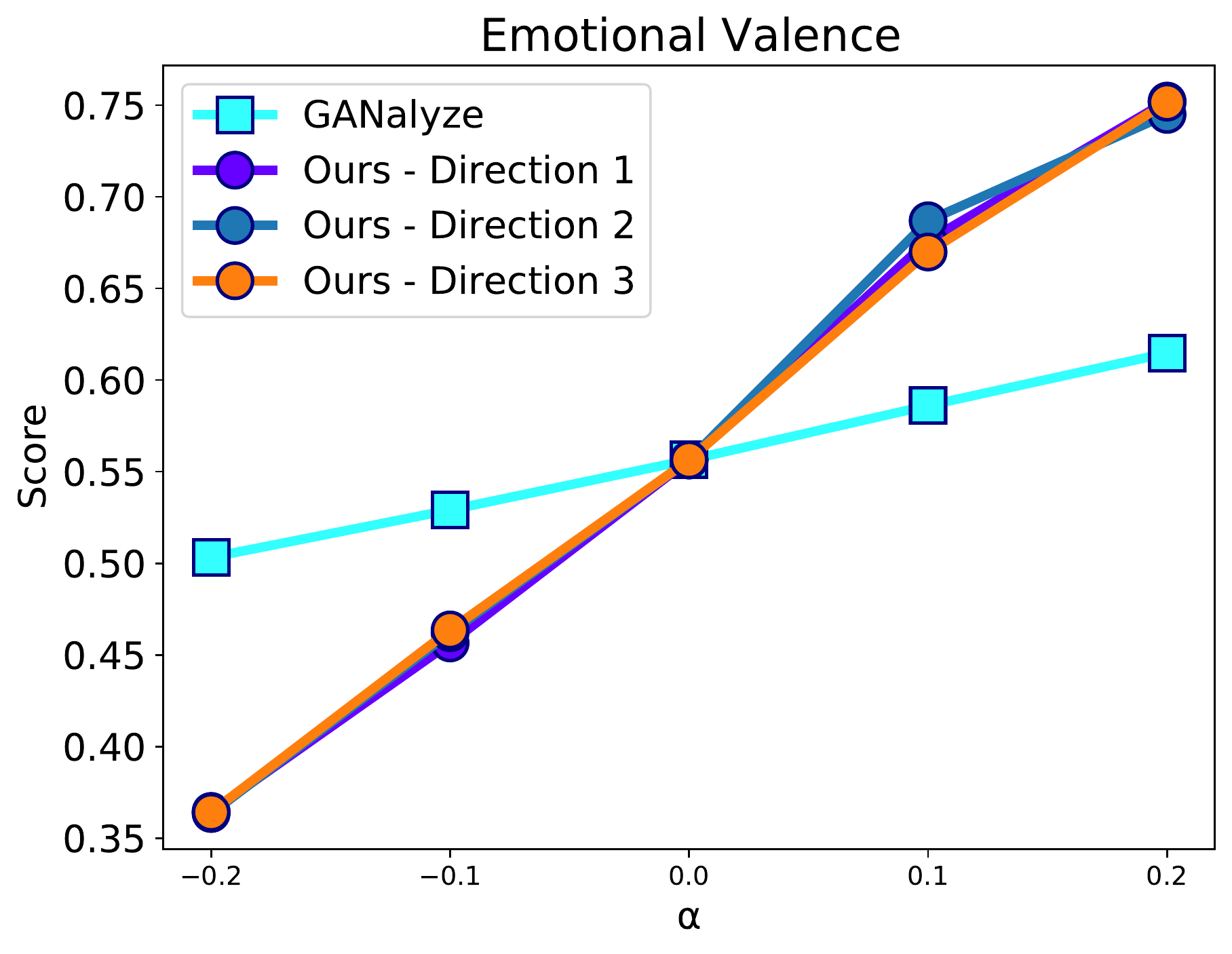}
  \includegraphics[width=0.32\linewidth]{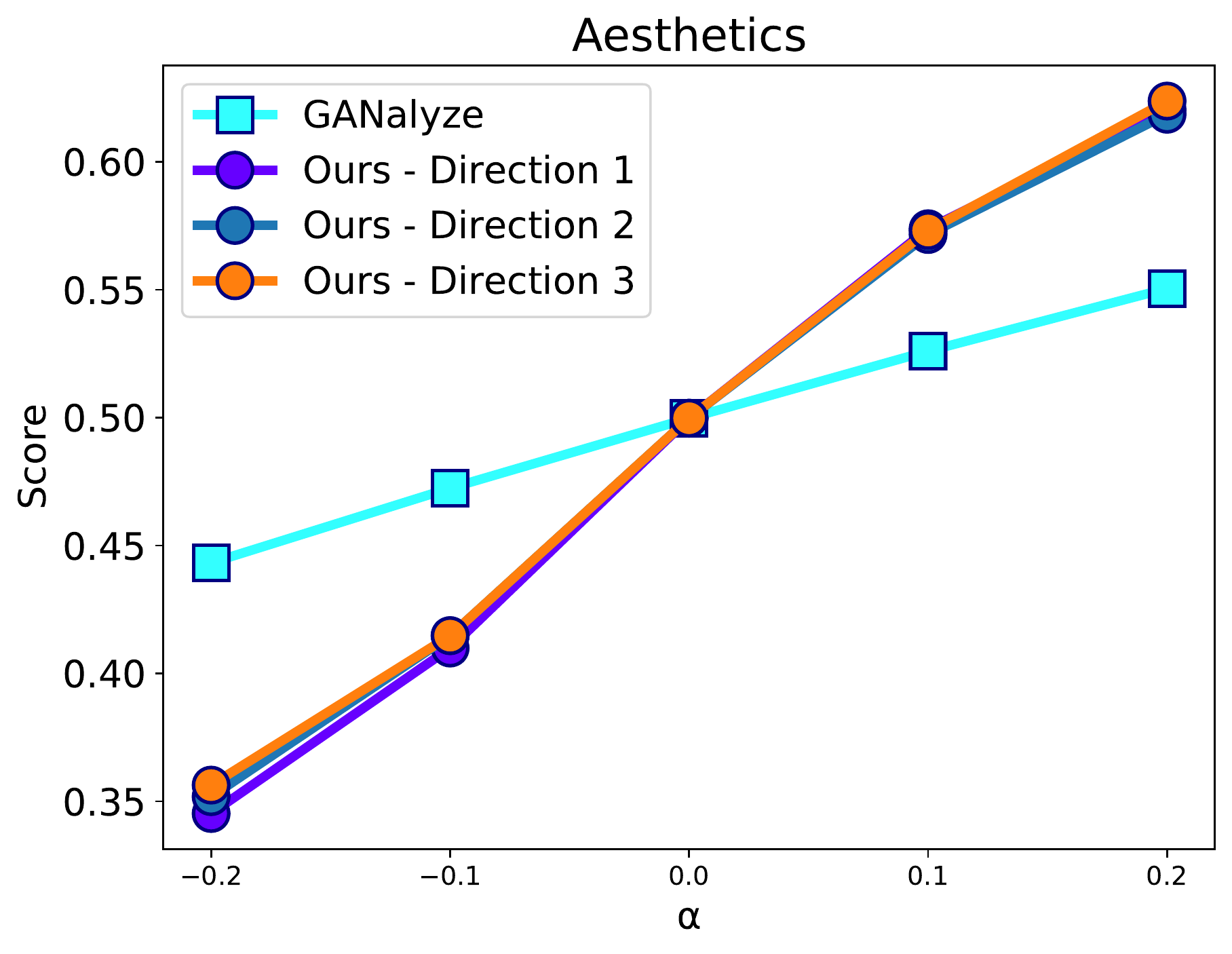}
        \end{tabular}

  \end{center}
  \caption{The effect of learning single and multiple directions on Memorability, Emotional Valence, and Aesthetics. Our method makes better manipulation compared to GANalyze \cite{goetschalckx2019ganalyze}. For each figure, the $x$-axis represents the $\alpha$ value and the $y$-axis represents the mean value obtained by the corresponding scoring function.}
  \label{fig:exp_cond}
\end{figure*}

\section{Methodology}
\label{sec:methodology}
In this work, we use a pre-trained GAN model with a generator $\mathcal{G}$. $\mathcal{G}$ is a mapping function that takes a noise vector $\mathbf{z} \in \mathbb{R}$ of dimension $d$ and a one-hot class vector $y \in \{0,1\}$ of size $1000$ as input and generates an image $\mathcal{G}(\mathbf{z},y)$ as output. The latent code $\mathbf{z}$ is drawn from a prior distribution $p(\mathbf{z})$, typically chosen to be  a Gaussian. We also assume that we have access to a scorer function $\mathcal{S}$ that evaluates the property of interest, and $\alpha$ is a scalar value representing the degree of manipulation we wish to achieve. 

Our goal is to learn $k$ diverse directions formalized by functions $F_1, F_2, \ldots F_k$ by minimizing the following optimization problem:

\begin{equation*}
  \argmin_{} \sum_{i=1}^k \mathcal{L}_{\text{COND}_i} + \lambda \mathcal{L}_{\text{DIV}}
\end{equation*}
 
where $\mathcal{L}_{\text{COND}}$ is the conditional loss, which is computed based on the input vector $\mathbf{z}$ and the direction $i$, and $\mathcal{L}_{\text{DIV}}$ is the regularization loss term that penalizes directions that are similar to each other. $\lambda$ is a hyperparameter that balances the editability and diversity of the learned directions. $\mathcal{L}_{\text{COND}}$ for a given direction $i$ is defined as:

\begin{equation*}
 \mathcal{L}_{\text{COND}_i} = \mathbb{E}_{\mathbf{z},\mathbf{y},\alpha} [ (\mathcal{S}(\mathcal{G}(F_i(\mathbf{z}, \alpha), \mathbf{y}))  -  (\mathcal{S}(\mathcal{G}(\mathbf{z}, \mathbf{y})) + \alpha))^2 ] 
\end{equation*}

where the first term represents the score of the modified image after applying the conditional edit function $F_i$ given parameters $\mathbf{z}$ and $\alpha$, and the second term simply represents the score of the original image $\mathcal{S}(\mathcal{G}(\mathbf{z},y))$ increased or decreased by $\alpha$. $F_{i}$ computes a conditional direction using direction index $i$ and a neural network $\mathbf{NN}$ that consists of two dense layers with a RELU nonlinearity in between:

$$F_i(\mathbf{z}, \alpha) = \mathbf{z} + \alpha \cdot \mathbf{NN}_i(\mathbf{z})$$
$\mathcal{L}_{\text{DIV}}$ acts as a regularizer and penalizes the directions that are similar to each other. We define $\mathcal{L}_{\text{DIV}}$ as follows:

\begin{equation*}
 \mathcal{L}_{\text{DIV}} = \sum_{i,j, i\neq j}^{k} \mathbf{F_{\text{SIM}}} (F_i(\mathbf{z}, \alpha), F_j(\mathbf{z}, \alpha))
\end{equation*}
 
where $\mathbf{F_{\text{SIM}}}$  computes the similarity between two input vectors:

\begin{equation*}
\mathbf{F_{\text{SIM}}} (F_i(\mathbf{z}, \alpha), F_j(\mathbf{z}, \alpha)) = \dfrac{ F_i(\mathbf{z}, \alpha) \cdot F_j(\mathbf{z}, \alpha) } {\lVert F_i(\mathbf{z}, \alpha) \rVert \cdot \lVert F_j(\mathbf{z}, \alpha) \rVert }  
\end{equation*}

In this work, we use the cosine similarity for $\mathbf{F_{\text{SIM}}}$, however any other similarity function that measures the similarity between two vectors can be used.

\section{Experiments}
\label{sec:experiments}
In this section, we first explain our experimental setup. We then present quantitative and qualitative experiments on Memorability, Emotional Valence, and Aesthetics. We compare our method to GANalyze \cite{goetschalckx2019ganalyze}, as it is the only competing method that manipulates cognitive properties. We conduct extensive experiments to demonstrate the superiority of our method using single and multiple directions, and perform statistical and visual analysis to illustrate the performance of our method. In addition, we investigate the diversity and image quality of the generated images, and explore image factors such as \textit{colorfulness, squareness} or \textit{object size} to understand how our method alters images to manipulate cognitive properties.

\subsection{Experimental Setup}
\label{sec:experimental_setup}
For the Memorability, Emotional Valence, and Aesthetics experiments, we use a batch size of 4, Adam optimizer and a learning rate of $2e-4$. Similar to \cite{goetschalckx2019ganalyze, jahanian2019steerability}, we use a pre-trained BigGAN \cite{BigGAN} model to find directions in latent space. We use the BigGAN-256 model, which generates $256\times256$ images\footnote{BigGAN-256: \url{https://tfhub.dev/deepmind/biggan-256/2}.}. Since the model was trained on ImageNet \cite{russakovsky2015imagenet}, the dimension of the class vector is 1000 for both generators while the latent dimension is $140$ for BigGAN-256 and $128$ for BigGAN-512. Note that our method is applicable to any GAN architecture such as StyleGAN \cite{StyleGAN}, but we use BigGAN in our experiments since it provides a large set of diverse classes.

For all cognitive properties, we randomly generate a training set with $200K$ $\mathbf{z}$ vectors from a standard normal distribution truncated between $[-2,2]$. We associate each sample $\mathbf{z}$ with a randomly drawn $\alpha$ value from a uniform distribution between $[-0.5, 0.5]$ and a randomly chosen class $y$. After training the model on this training set, we use the obtained directions on test data consisting of randomly drawn $\mathbf{z}$ vectors.  
\begin{figure*}[!ht]
\vskip 0.2in
\qquad  \quad  \, \,  $\alpha$- -  \qquad \, \, $\alpha$=0 \quad \,  \qquad $\alpha$++ \qquad \, \quad  \, $\alpha$- - \; \; \qquad $\alpha$=0 \; \quad \qquad $\alpha$++ \qquad  \quad  \, $\alpha$- - \quad \, \qquad $\alpha$=0  \quad \qquad $\alpha$++
\vspace{-0.15cm} 
\begin{center}
 \begin{turn}{90} \quad \, Ours \qquad GANalyze \quad \; \, Ours \quad \; \, GANalyze \end{turn}
\includegraphics[width=0.66\columnwidth]{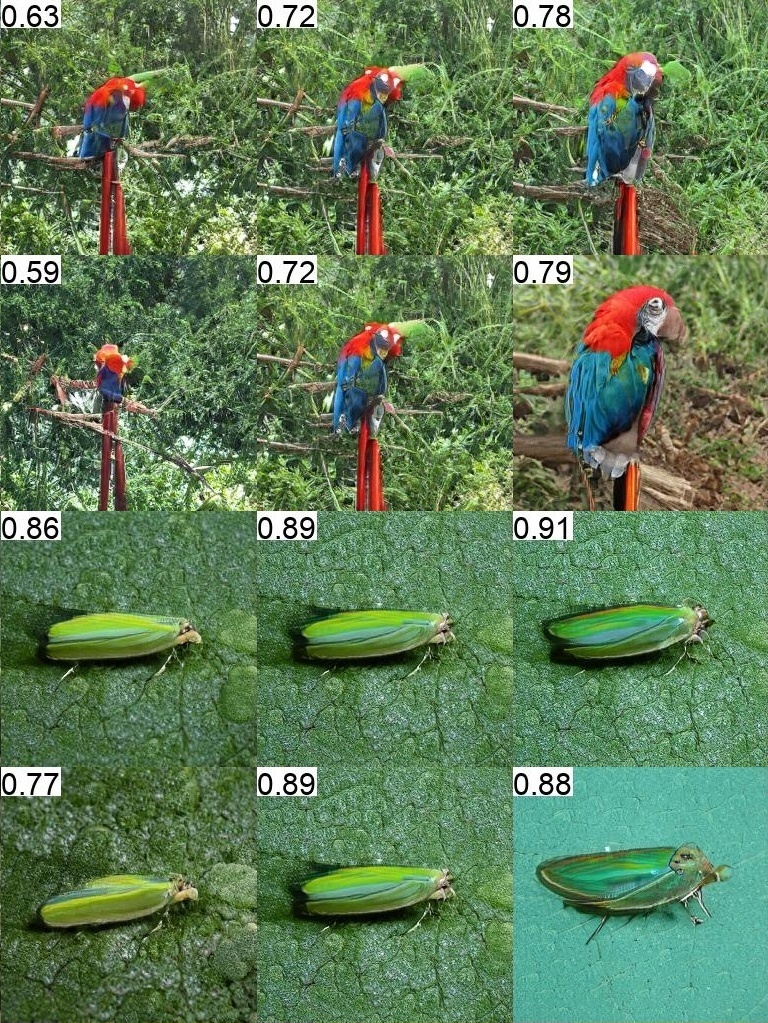}
\includegraphics[width=0.66\columnwidth]{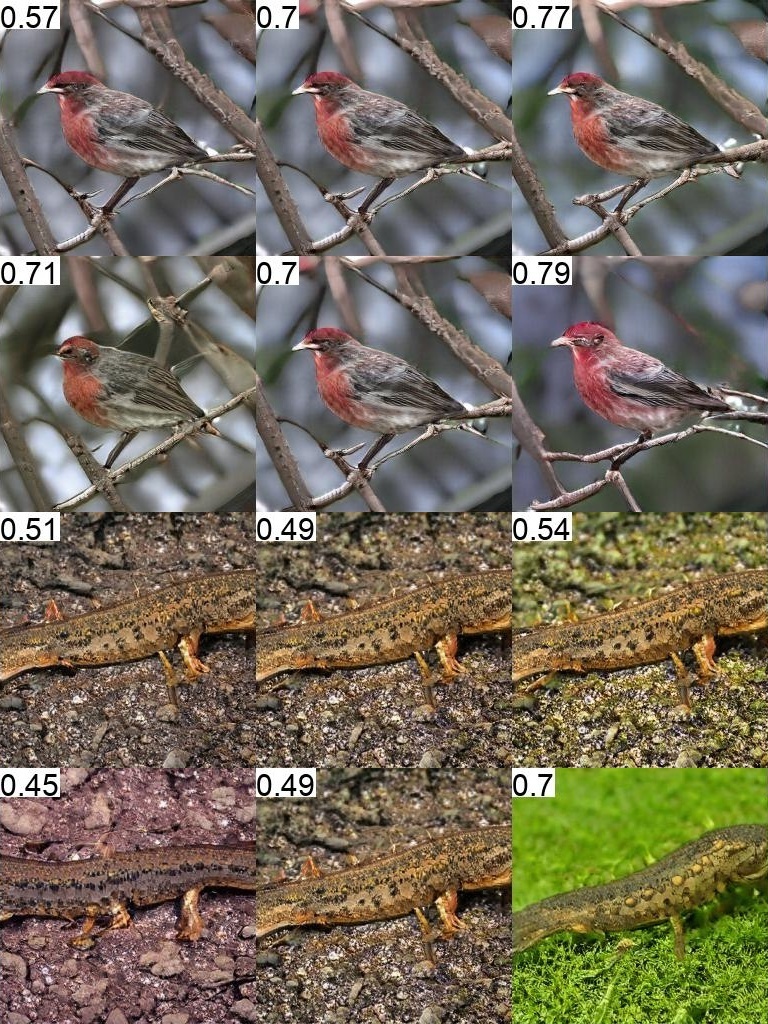}
\includegraphics[width=0.66\columnwidth]{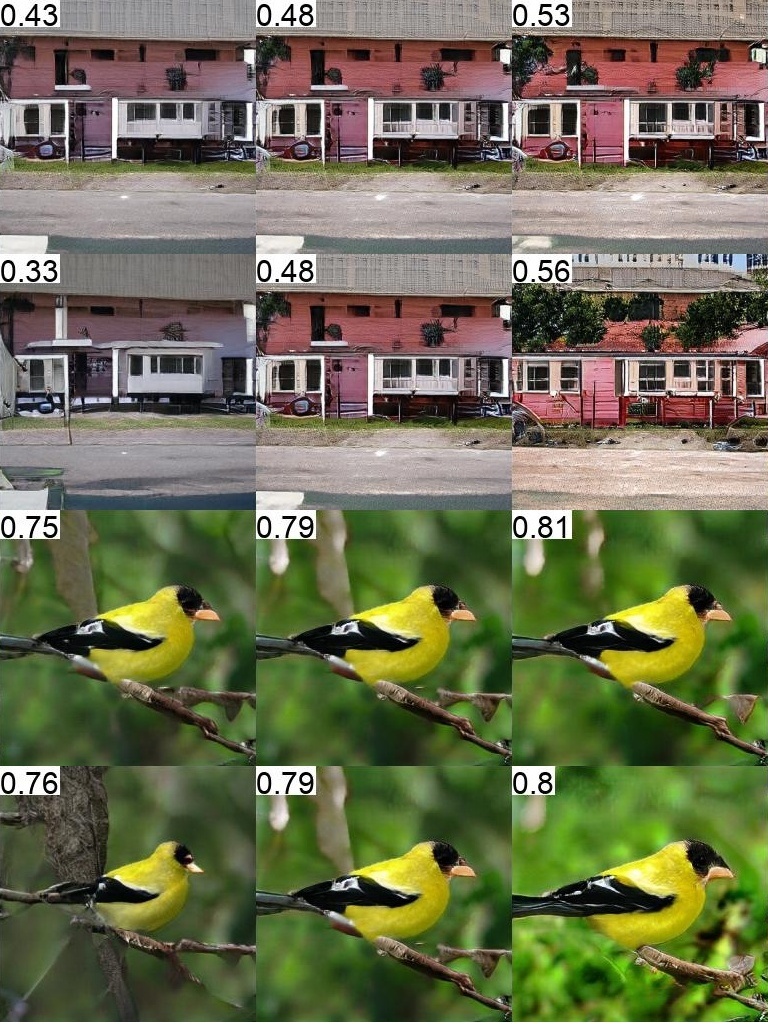}
\hspace{1.2cm} Memorability \hspace{3.3cm} Emotional Valence \hspace{3.3cm} Aesthetics
\caption{Examples of image manipulations for Memorability, Emotional Valence  and Aesthetics for our method and GANalyze \cite{goetschalckx2019ganalyze}. $\alpha=0$ denotes the input image, while $\alpha$- - and $\alpha$++ shift the latent code towards negative and positive directions, respectively. }
\label{fig:single_outputs}
\end{center}
\vskip -0.2in
\end{figure*}

\begin{figure*}[!ht]
\vskip 0.2in
\qquad  \quad  \, \,  $\alpha$- -  \qquad \, \, $\alpha$=0 \quad \,  \qquad $\alpha$++ \qquad \, \quad  \, $\alpha$- - \; \; \qquad $\alpha$=0 \; \quad \qquad $\alpha$++ \qquad  \quad  \, $\alpha$- - \quad \, \qquad $\alpha$=0  \quad \qquad $\alpha$++
\vspace{-0.15cm} 
\begin{center}
\begin{turn}{90} \quad Ours (1) \quad \, Ours (2) \quad \, Ours (3) \; \; GANalyze \quad \, Ours (1) \quad \, Ours (2) \quad \, Ours (3) \quad \; GANalyze \end{turn}
\includegraphics[width=0.66\columnwidth]{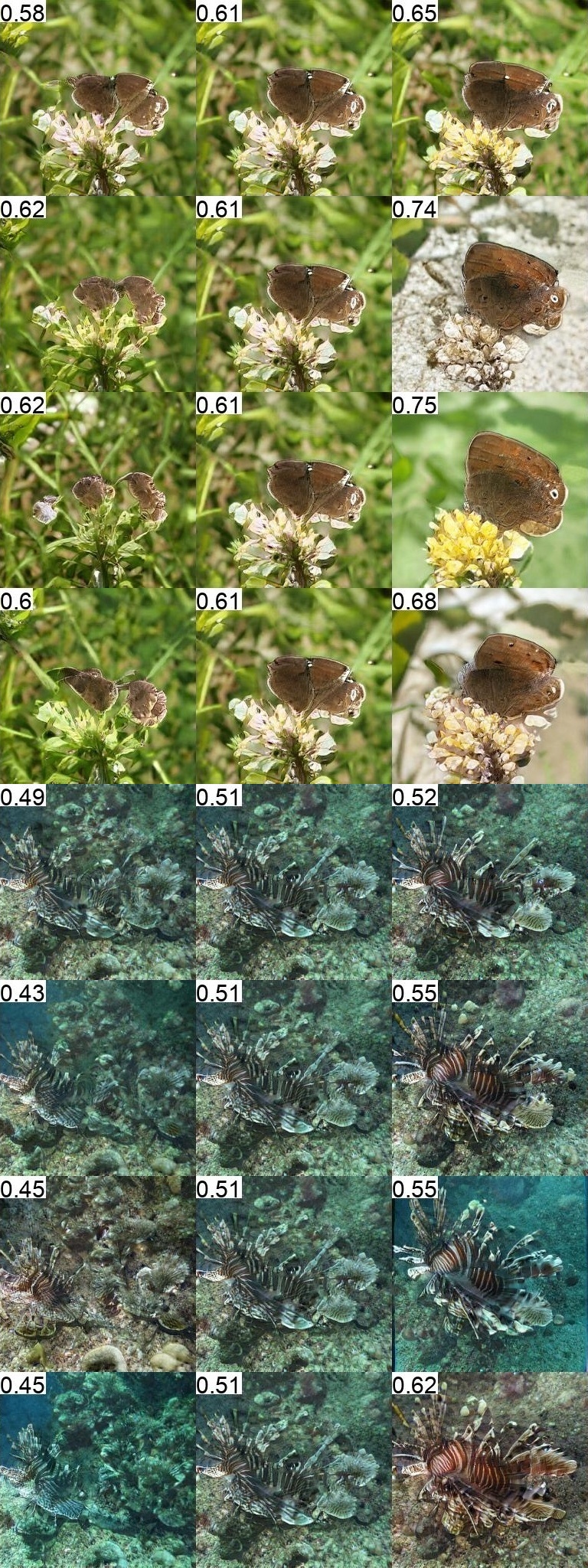}
\includegraphics[width=0.66\columnwidth]{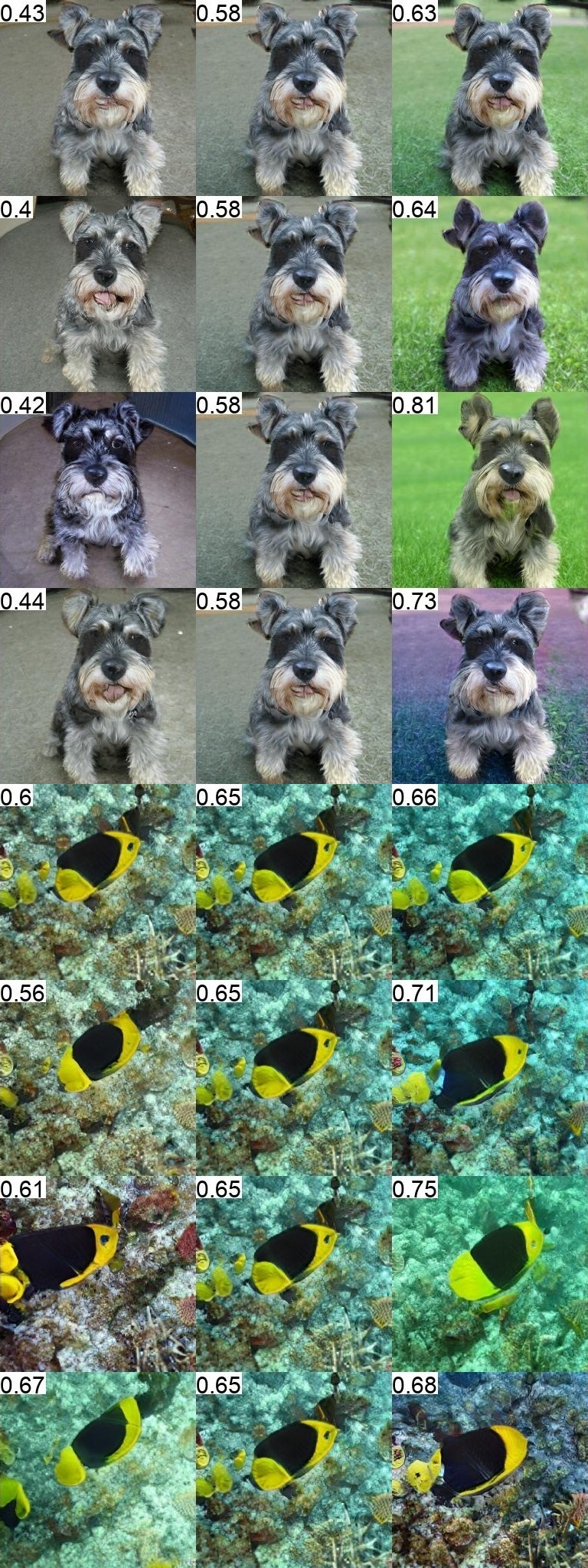}
\includegraphics[width=0.66\columnwidth]{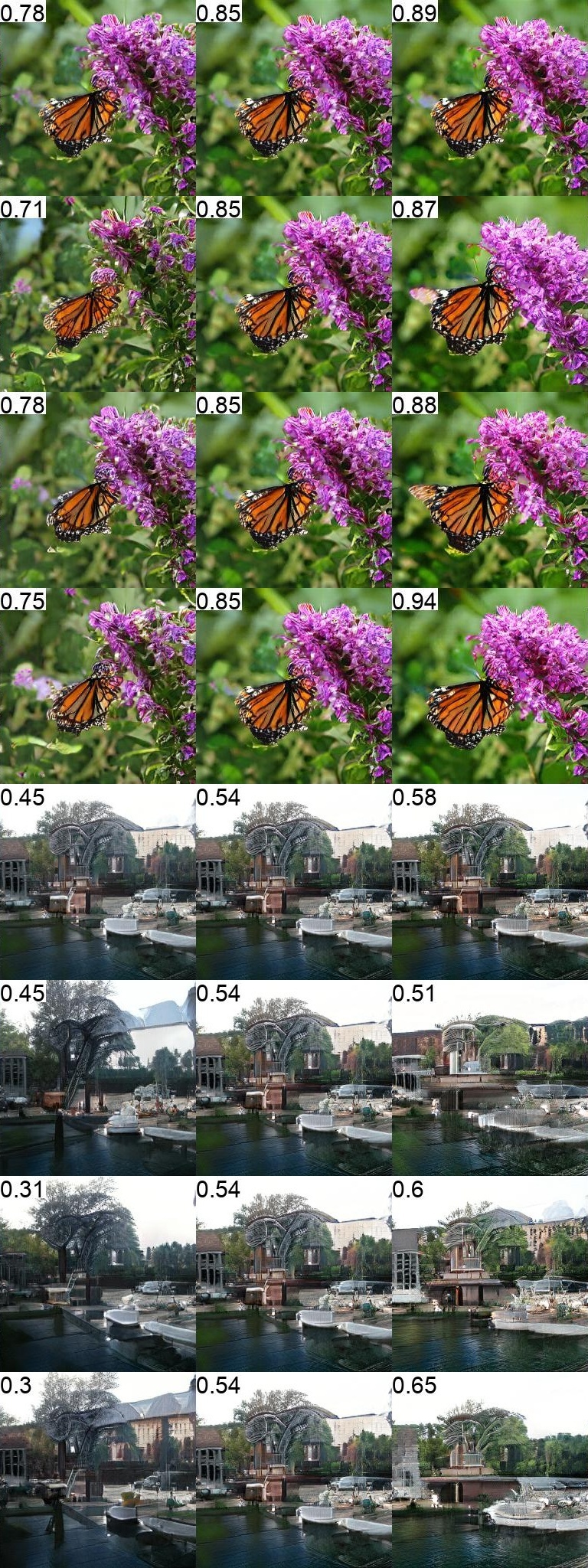}

\hspace{0.2cm} Memorability \hspace{3.3cm} Emotional Valence \hspace{3.3cm} Aesthetics

\caption{Examples of image manipulations for Memorability, Emotional Valence and Aesthetics for multiple directions $k=3$ using our method. $\alpha=0$ denotes the input image, while $\alpha$- - and $\alpha$++ shift the latent code towards negative and positive directions, respectively.}
\label{fig:single_outputs_multi}
\end{center}
\vskip -0.2in
\end{figure*}

\begin{figure*}[ht]
\vskip 0.2in
\qquad \quad \; \, Redness \quad \; \;  Colorfulness \quad \; \,  Brightness  \qquad \;  Entropy \qquad \:  Squareness \quad \, \,   Centeredness \quad \; \;    Object Size
\vspace{-0.3cm} 
\begin{center}
\begin{turn}{90} \quad Memnet \end{turn}
 \includegraphics[width=0.13\linewidth]{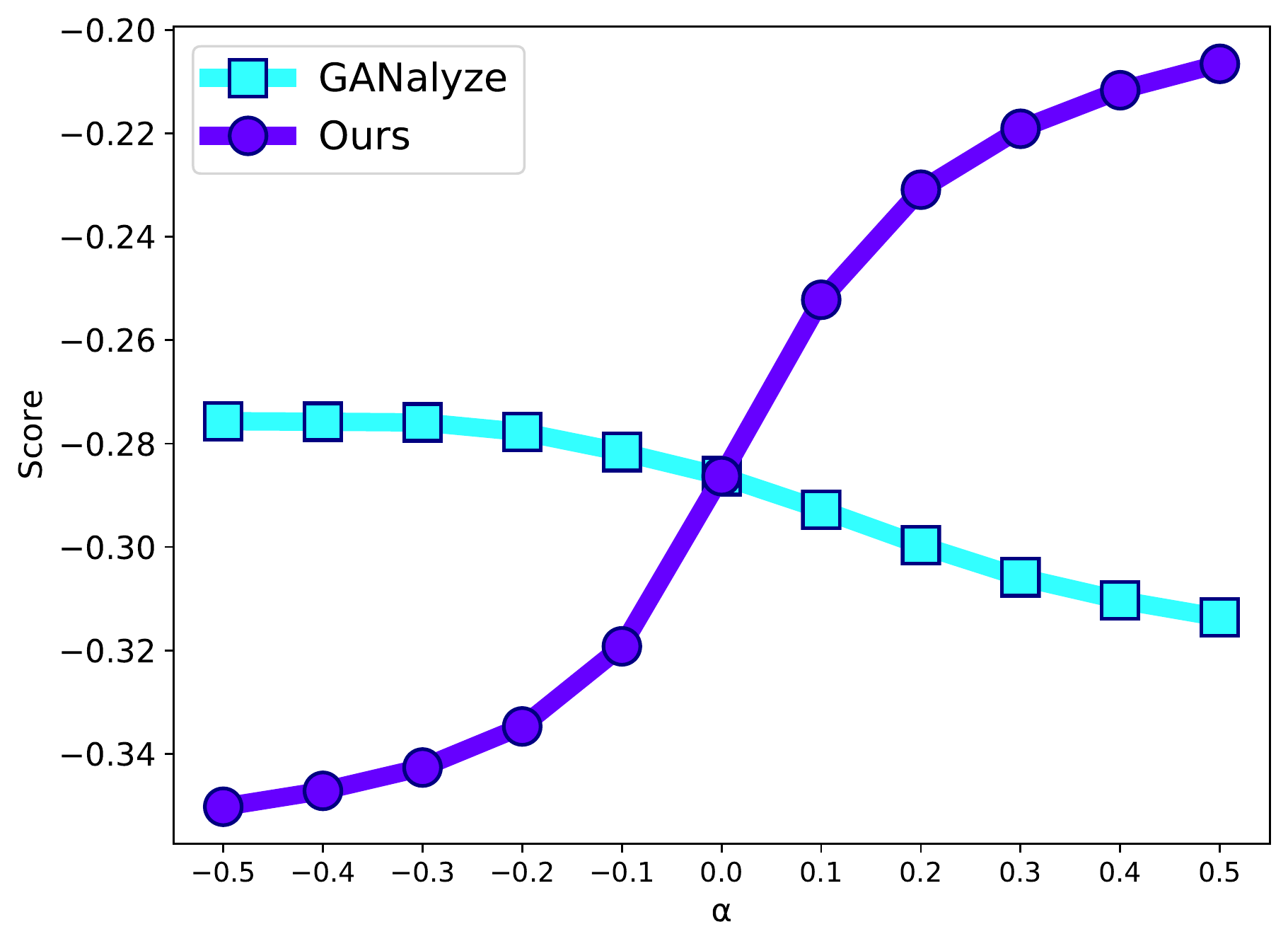} 
    \includegraphics[width=0.13\linewidth]{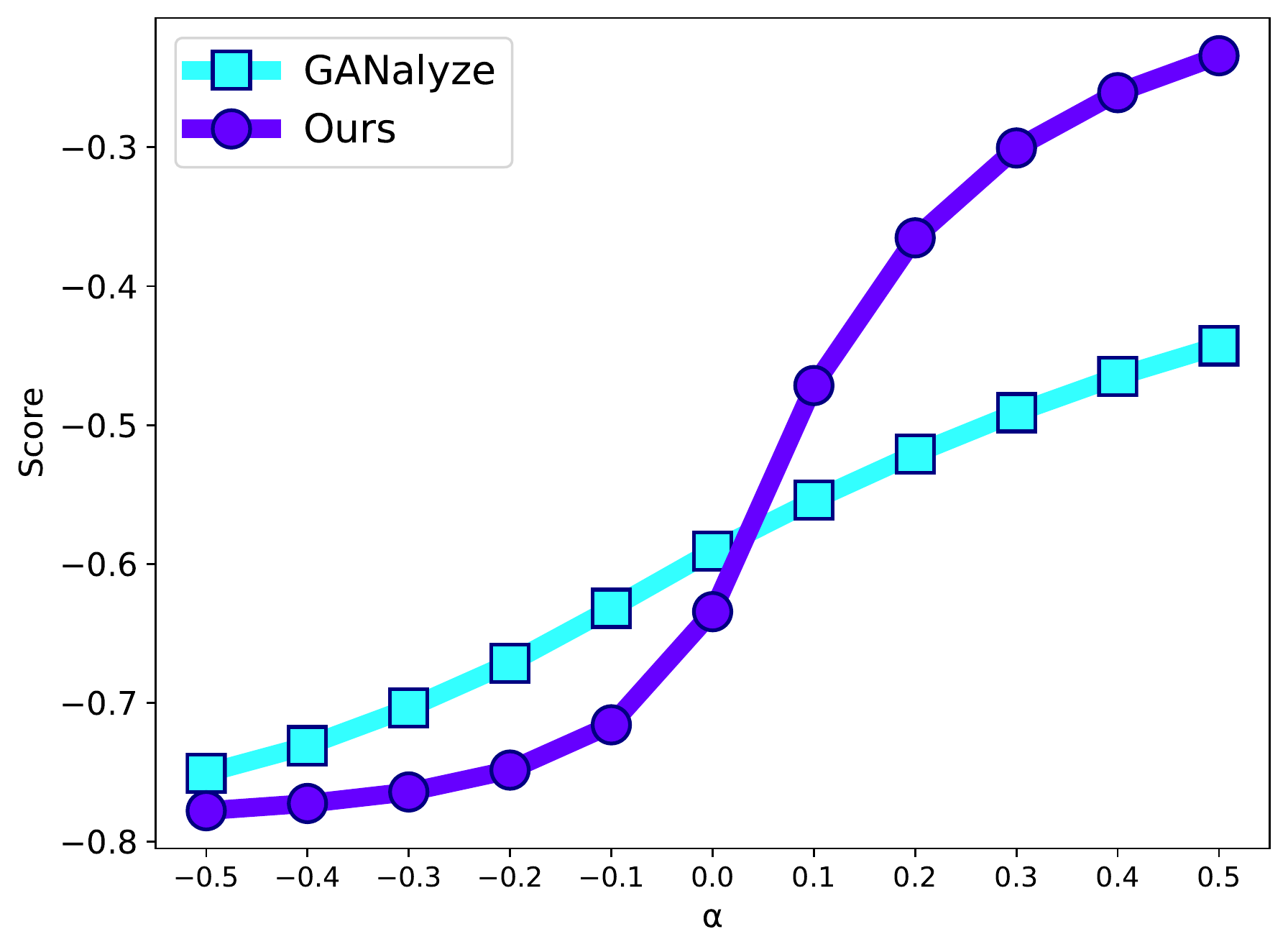} 
  \includegraphics[width=0.13\linewidth]{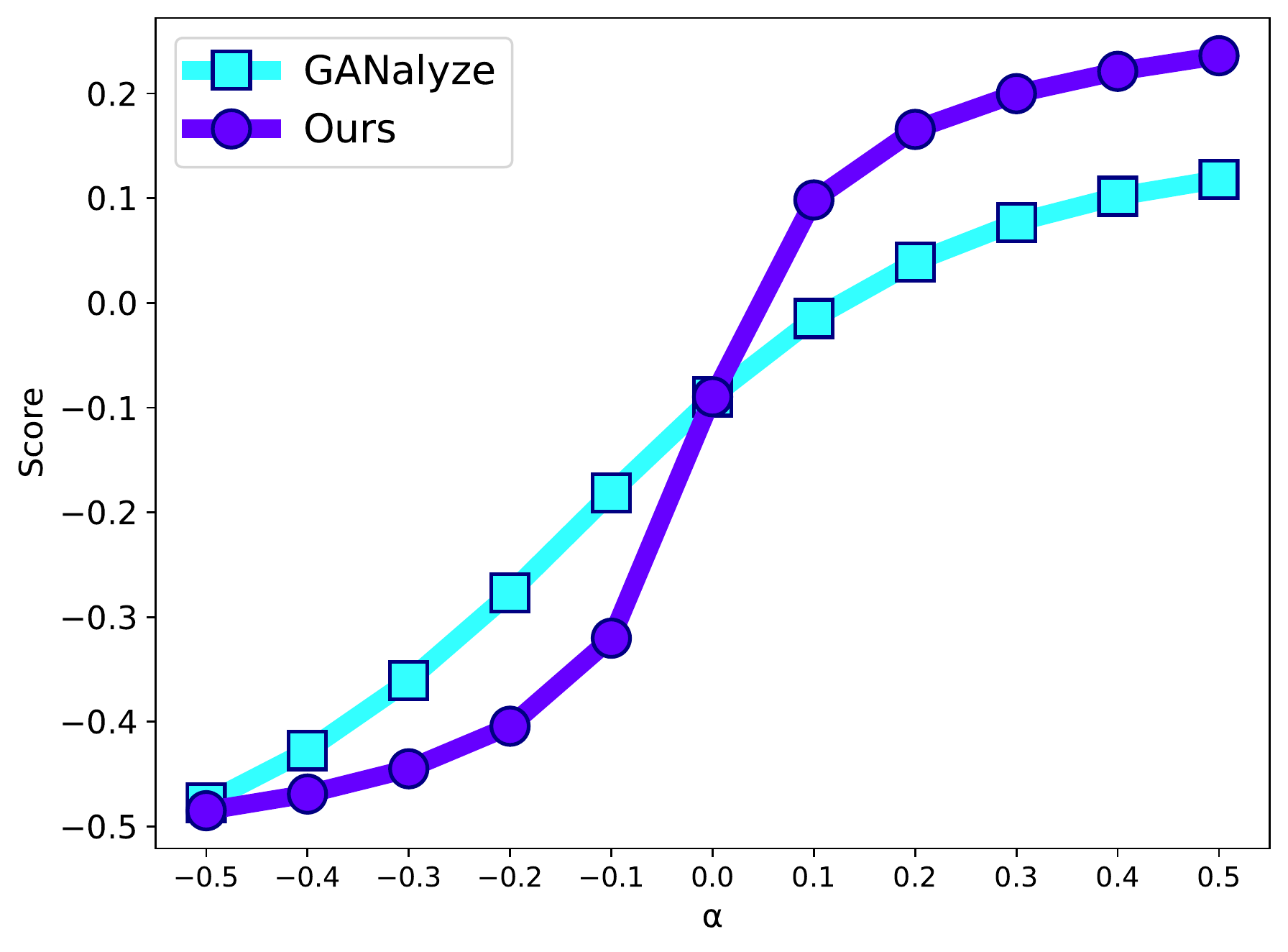} 
  \includegraphics[width=0.13\linewidth]{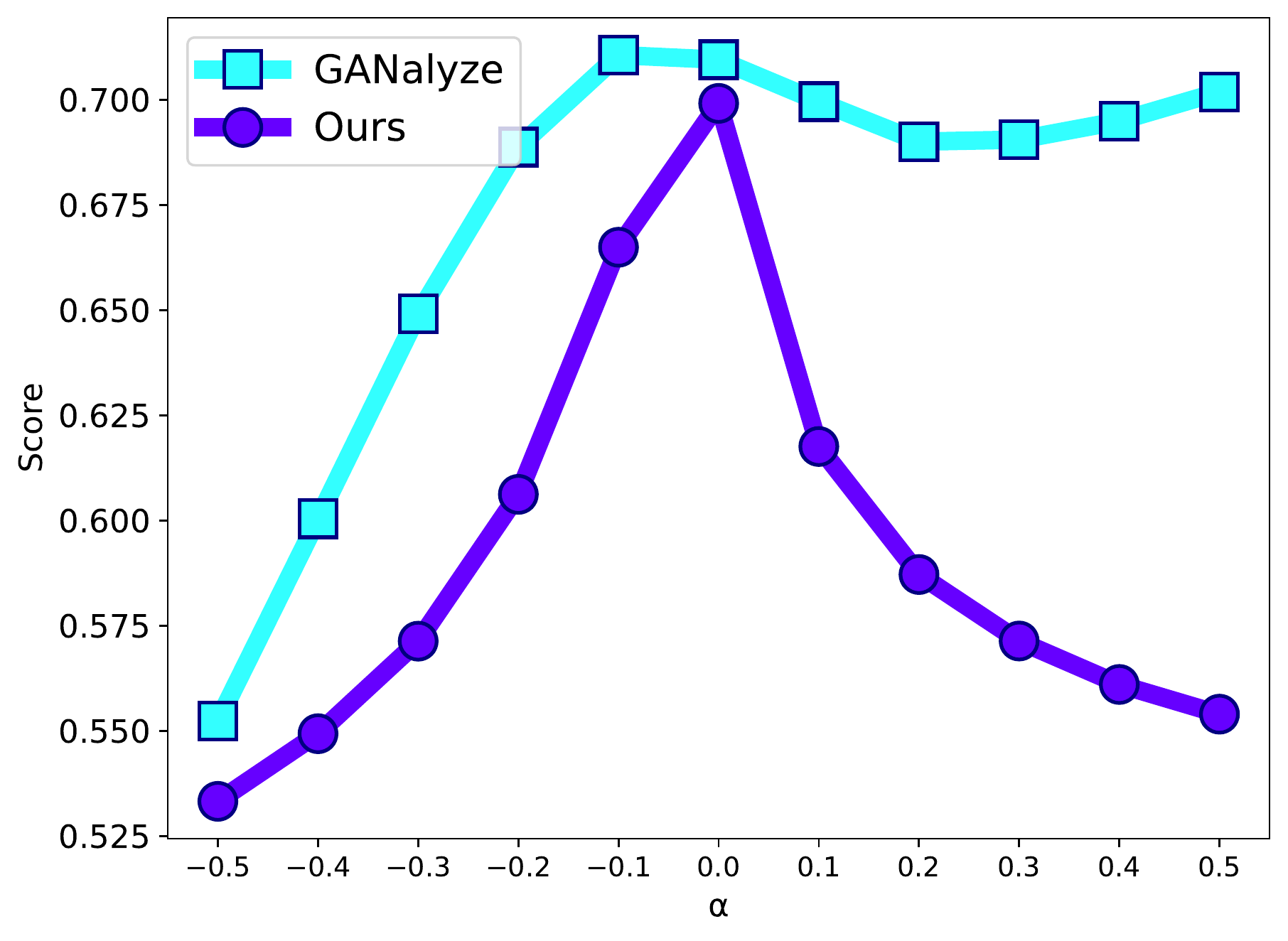} 
  \includegraphics[width=0.13\linewidth]{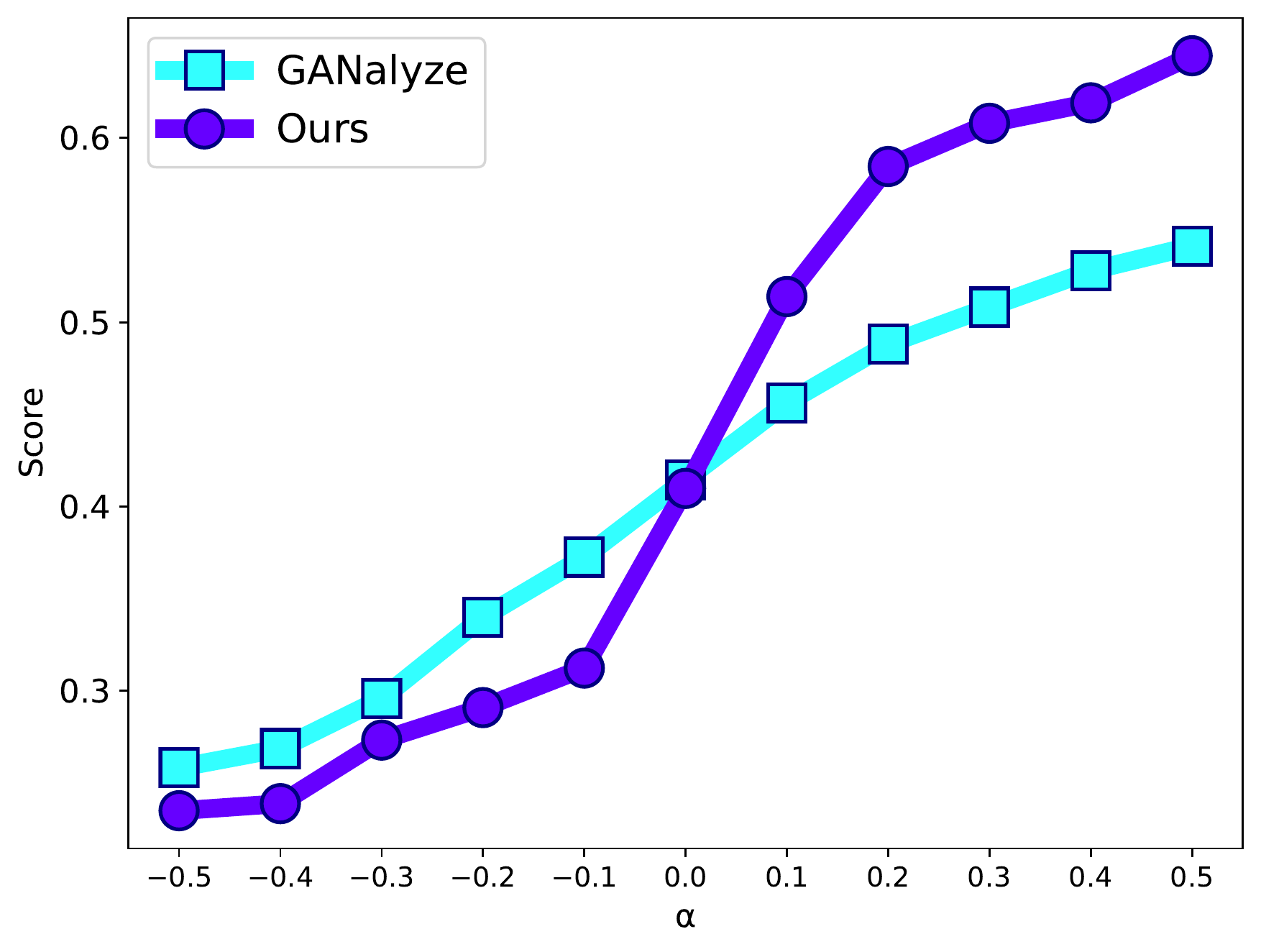} 
  \includegraphics[width=0.13\linewidth]{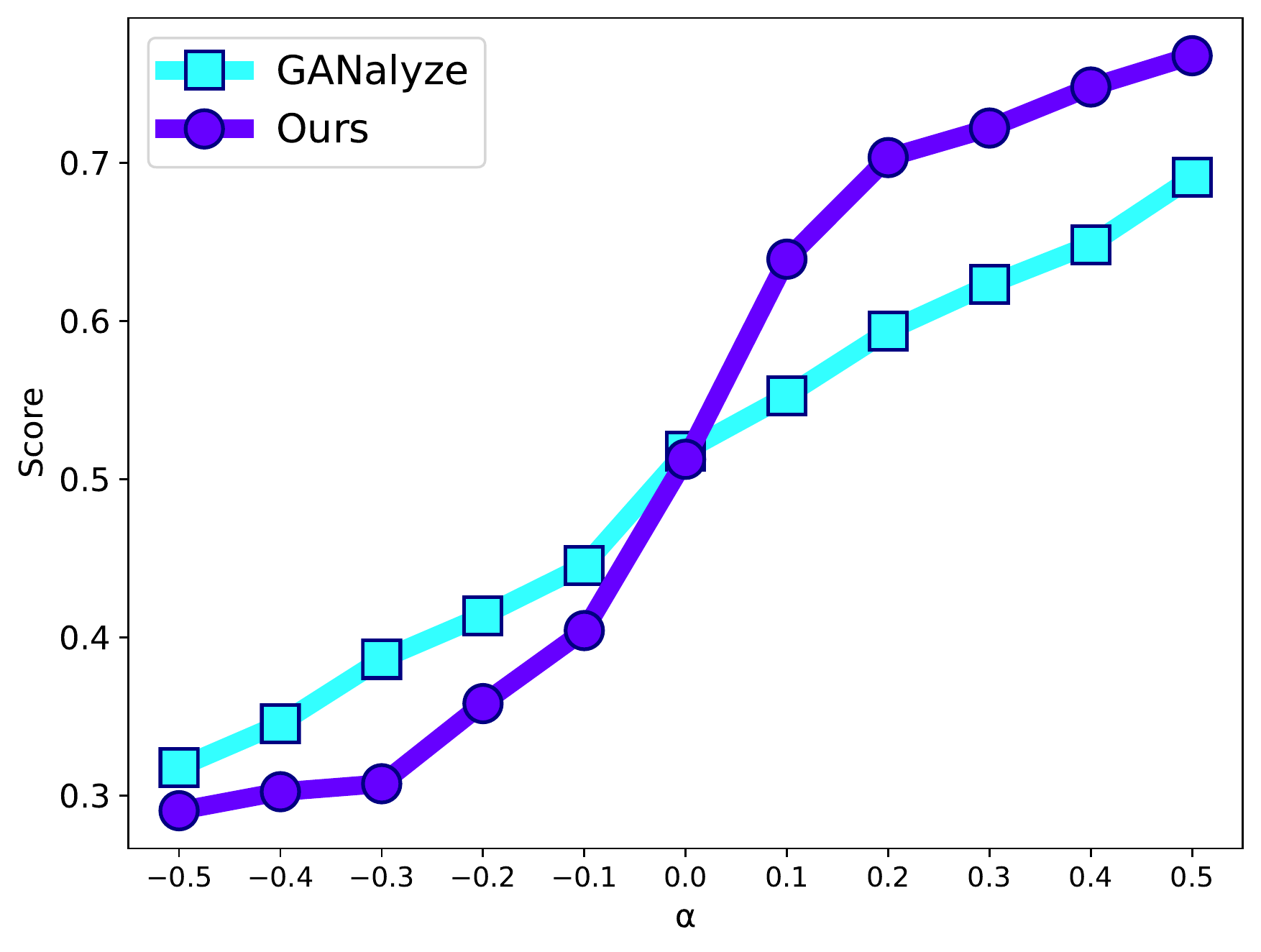} 
  \includegraphics[width=0.13\linewidth]{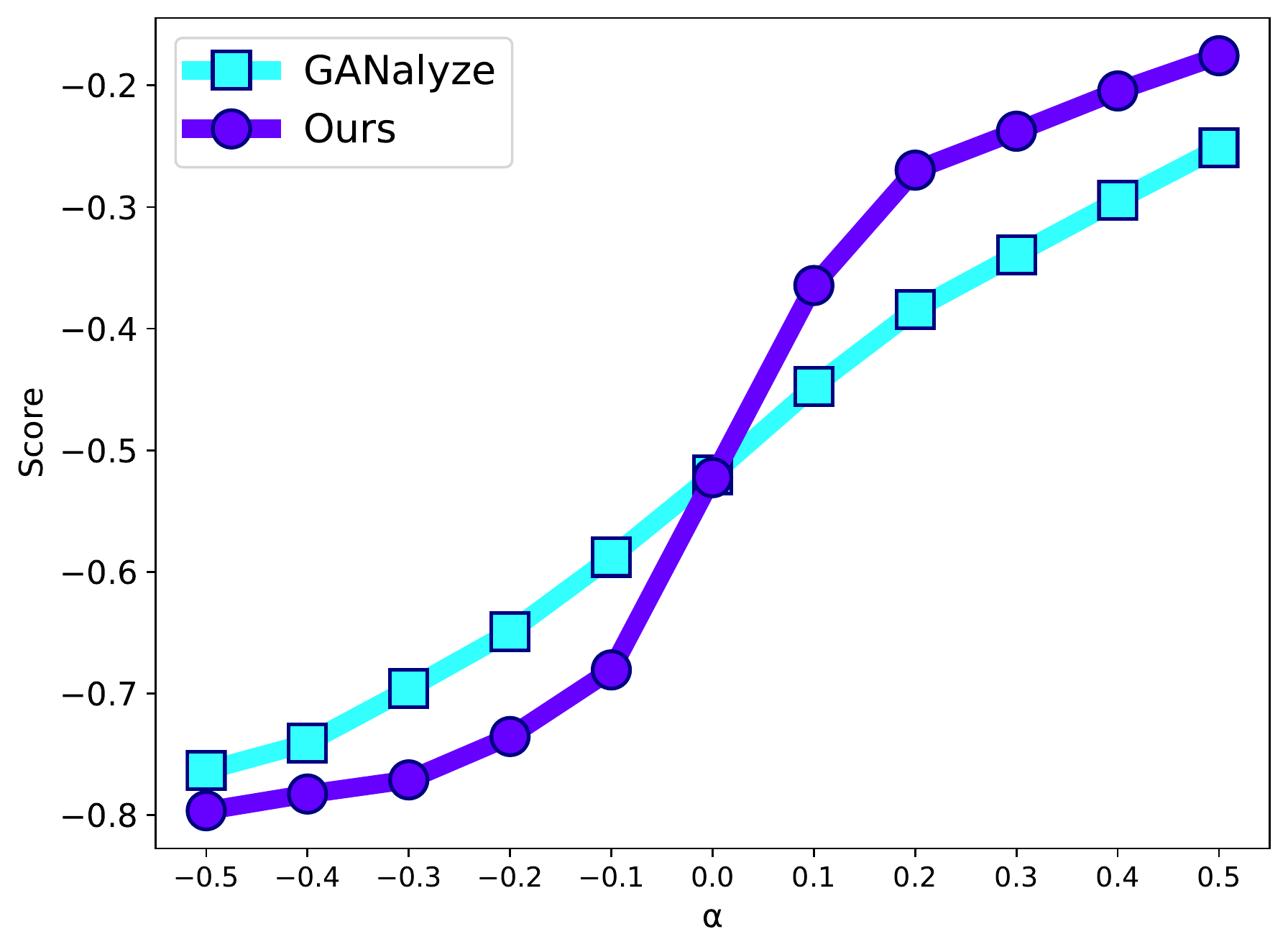} 
  
\begin{turn}{90} \quad Emonet \end{turn}
\includegraphics[width=0.13\linewidth]{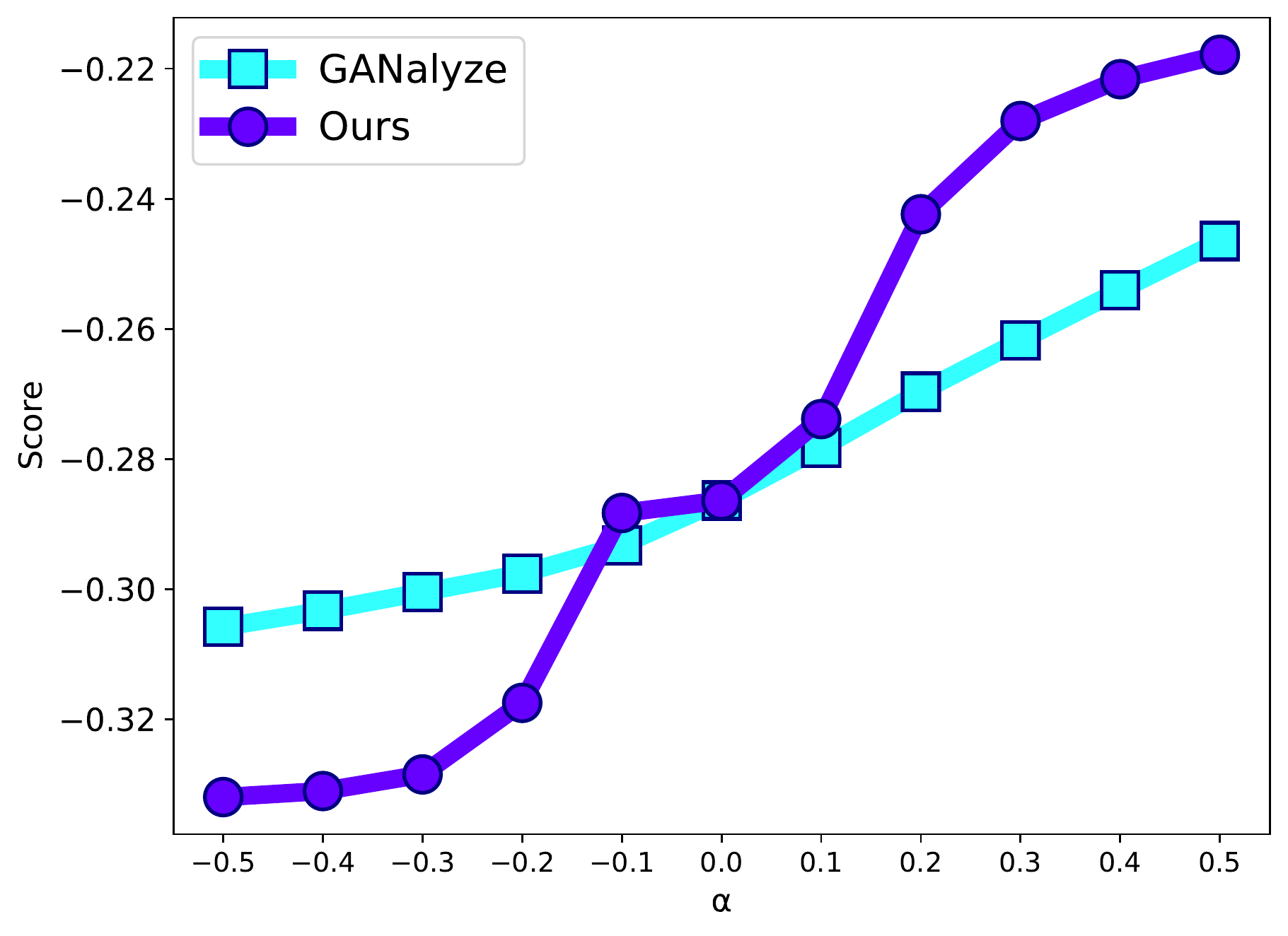} 
\includegraphics[width=0.13\linewidth]{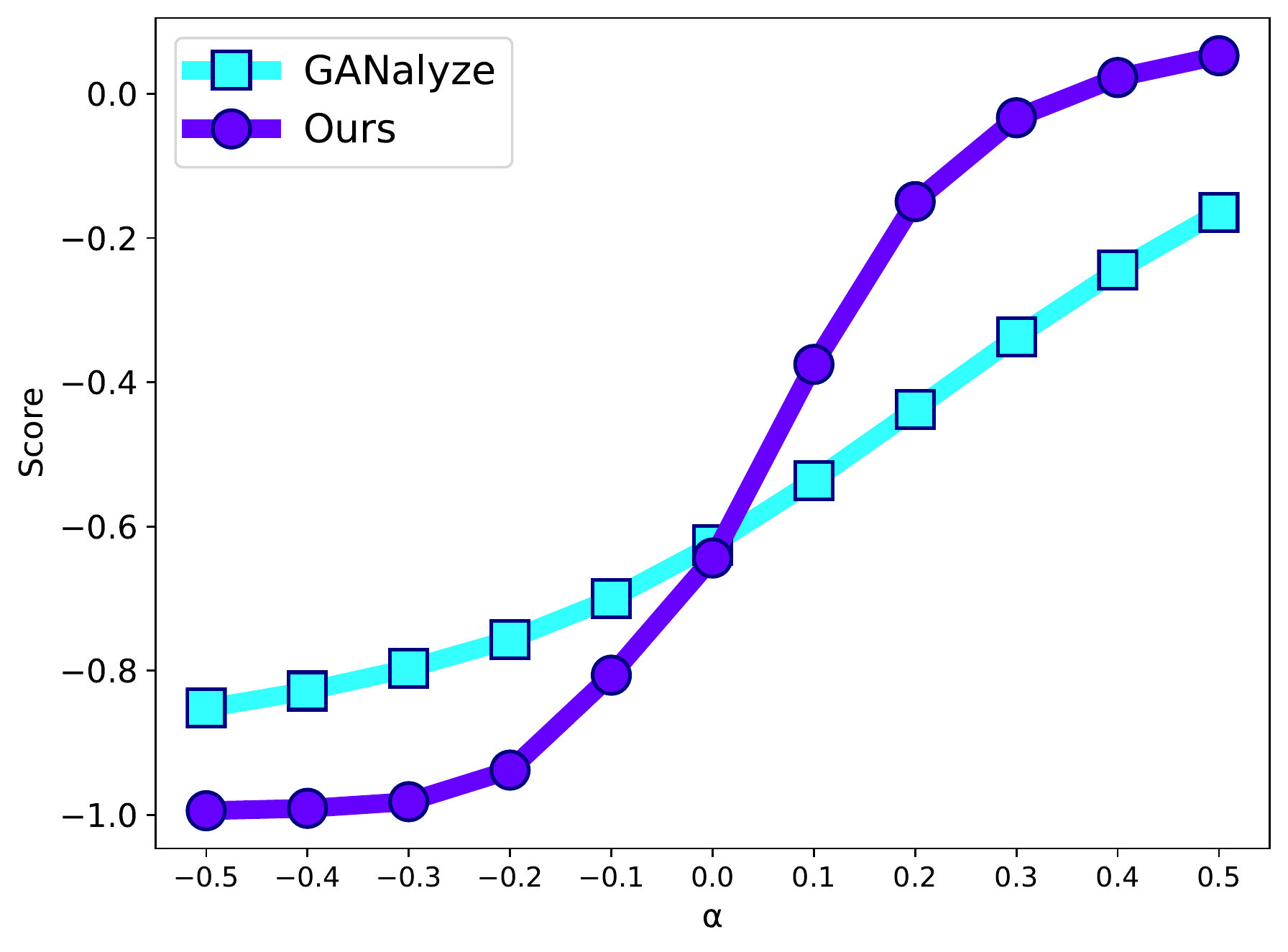}  
\includegraphics[width=0.13\linewidth]{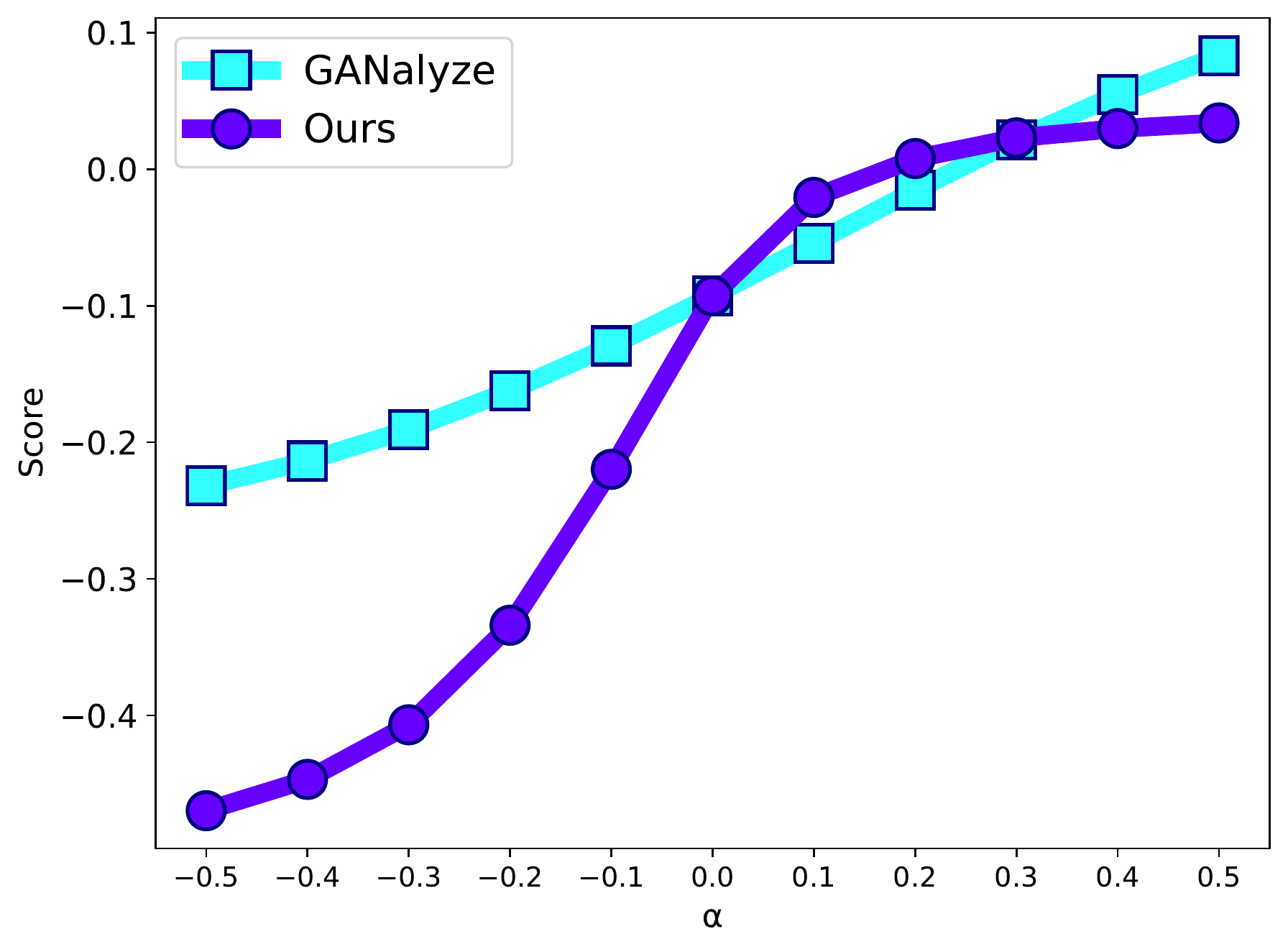}   
\includegraphics[width=0.13\linewidth]{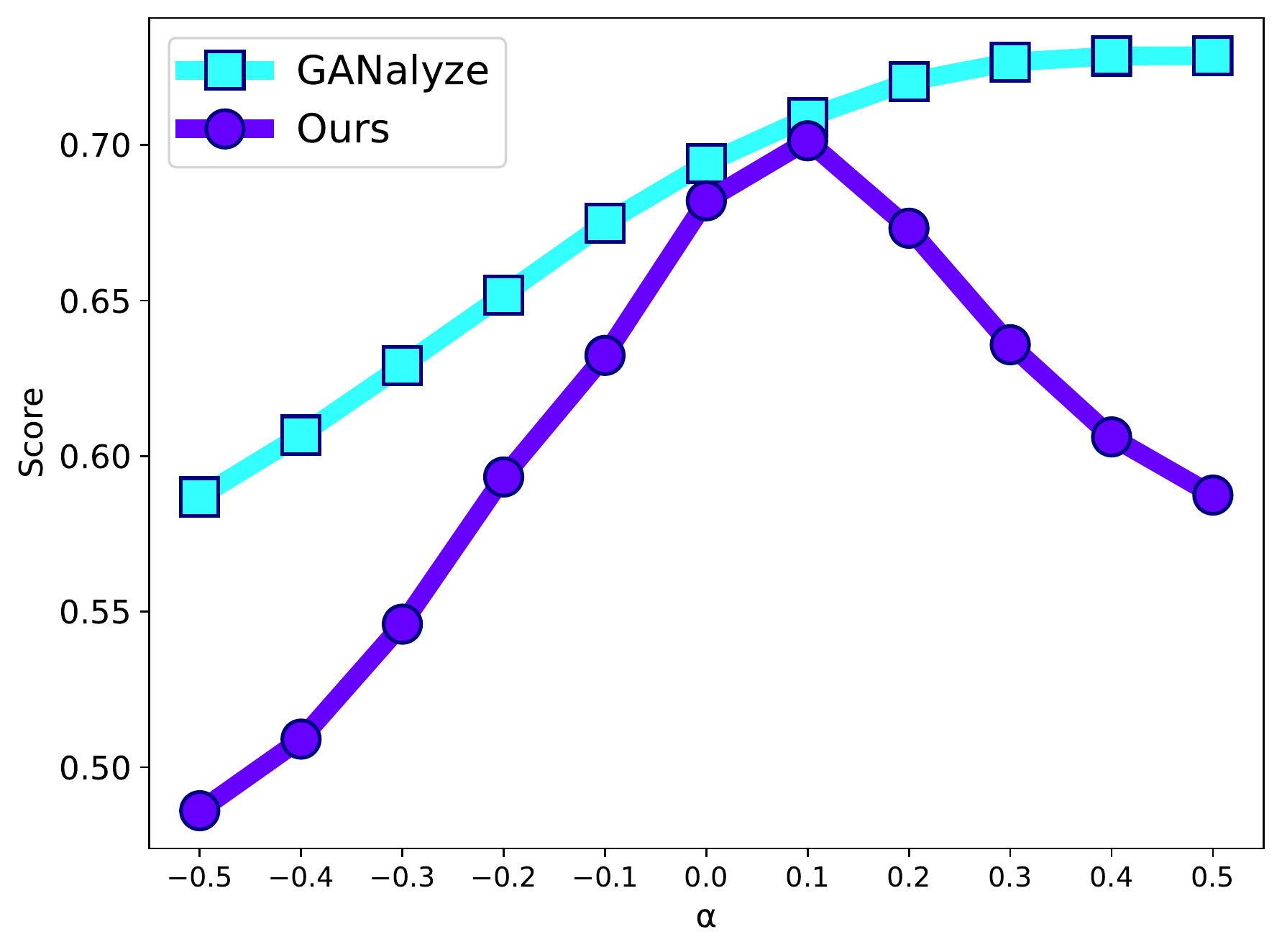} 
\includegraphics[width=0.13\linewidth]{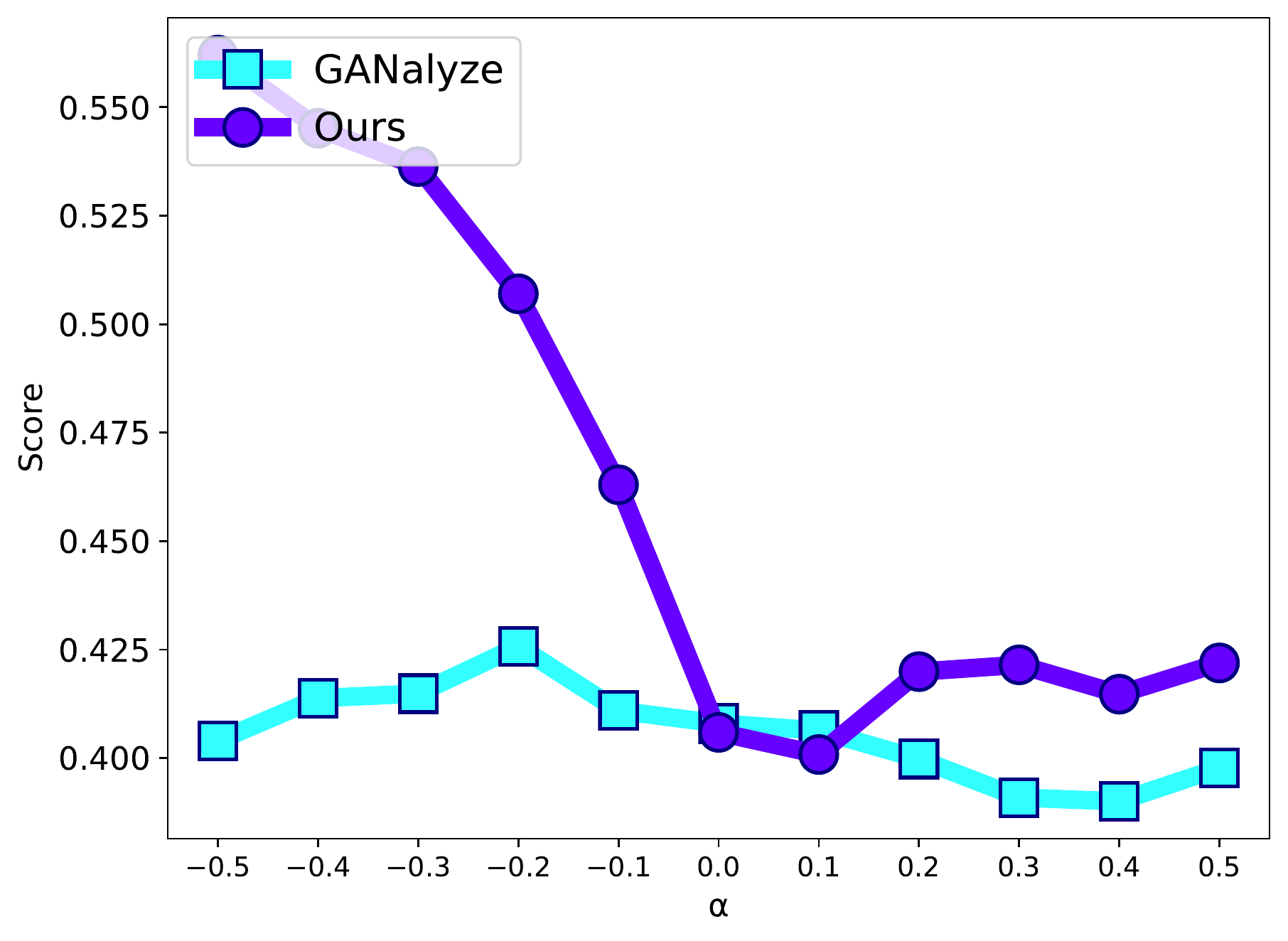} 
    \includegraphics[width=0.13\linewidth]{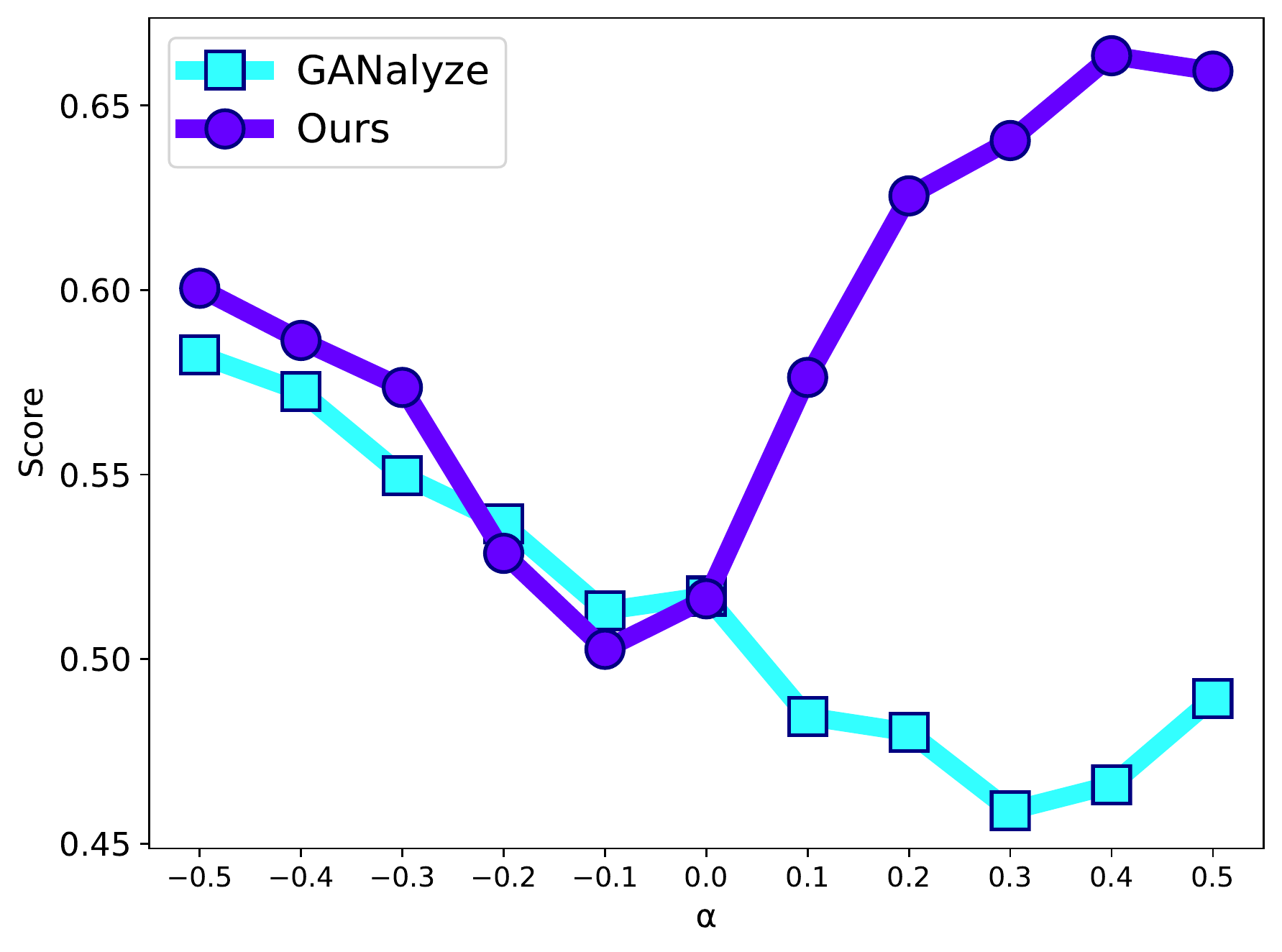} 
    \includegraphics[width=0.13\linewidth]{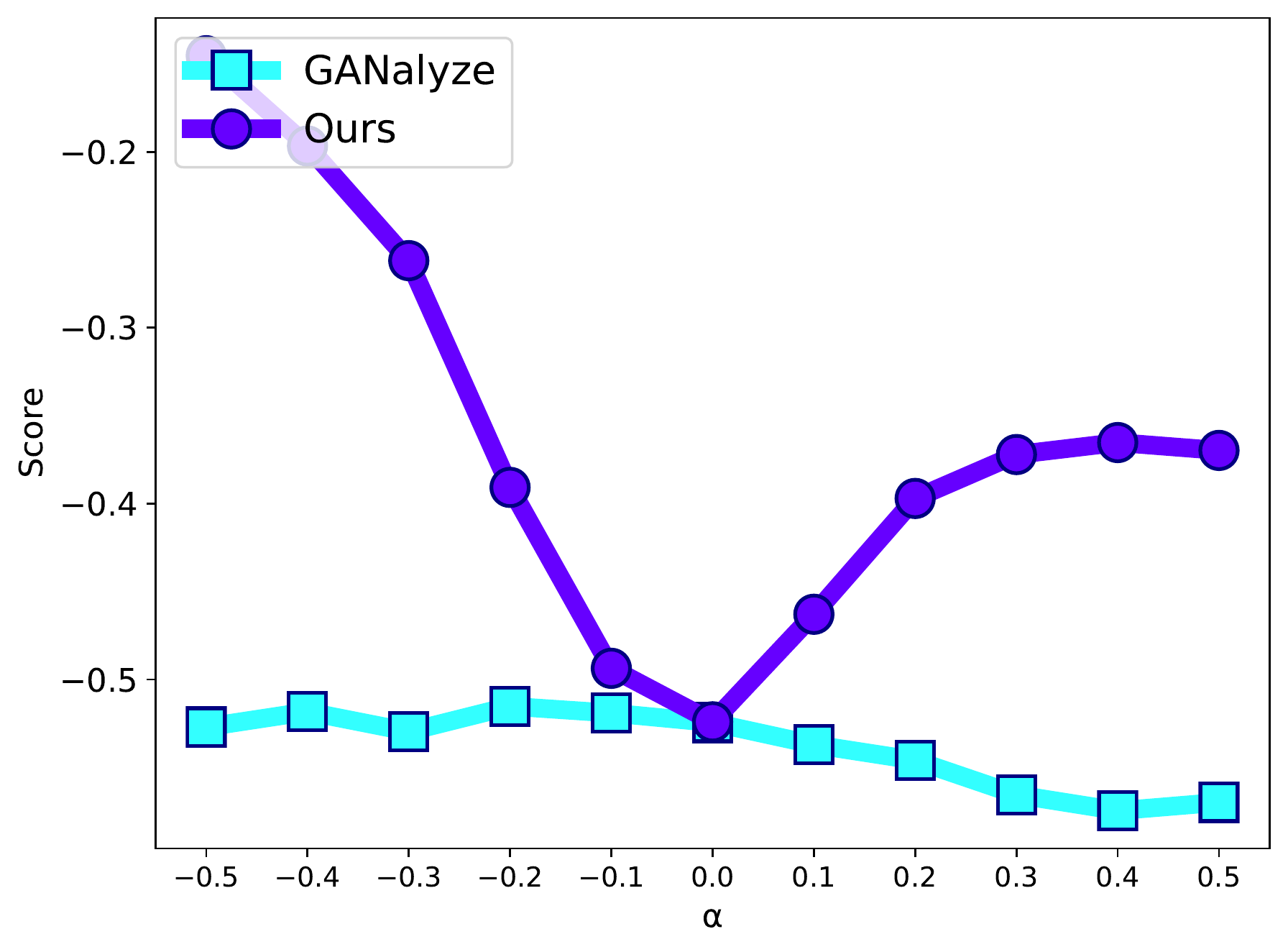}

\begin{turn}{90} \, Aesthetics \end{turn}
   \includegraphics[width=0.13\linewidth]{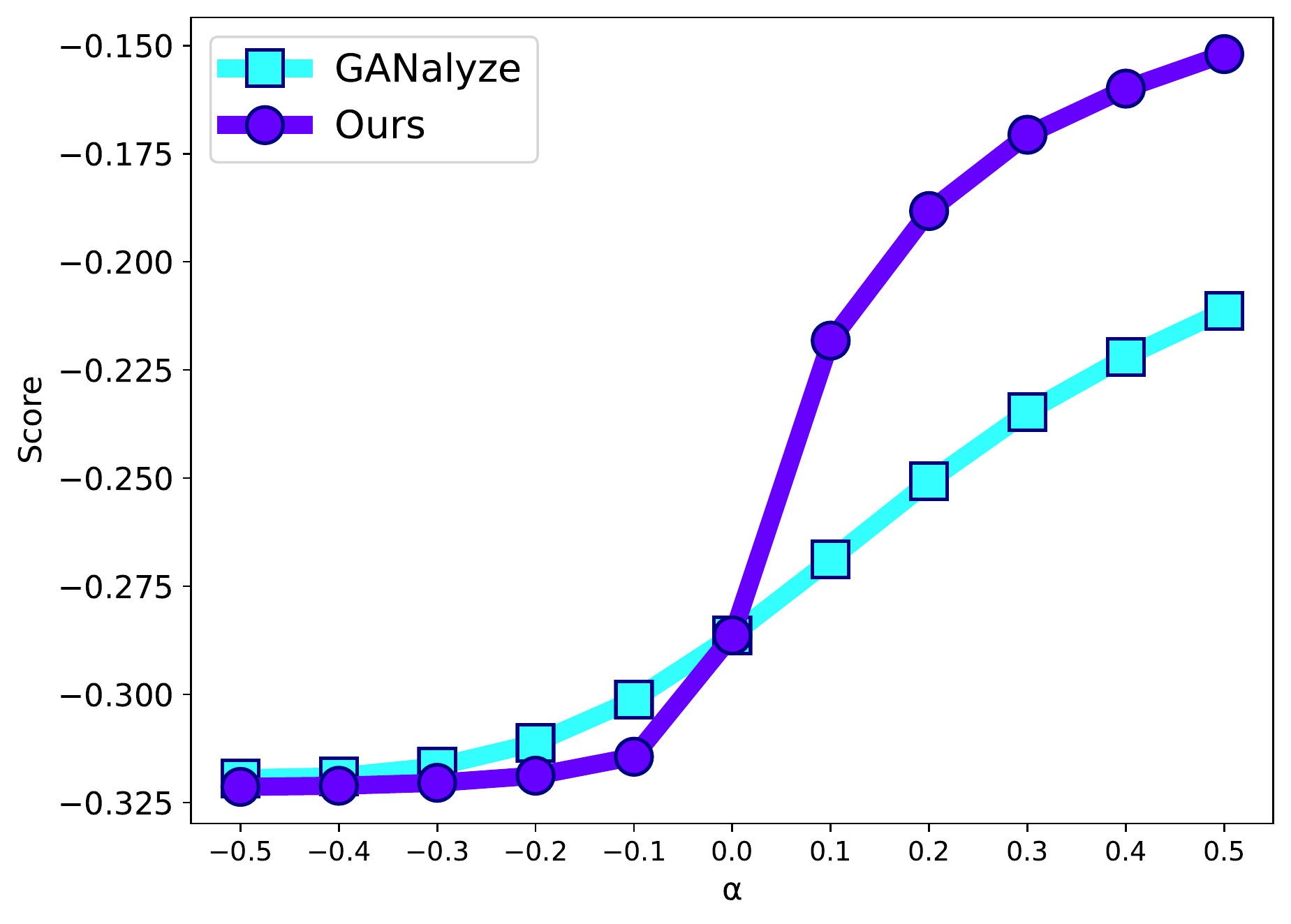} 
  \includegraphics[width=0.13\linewidth]{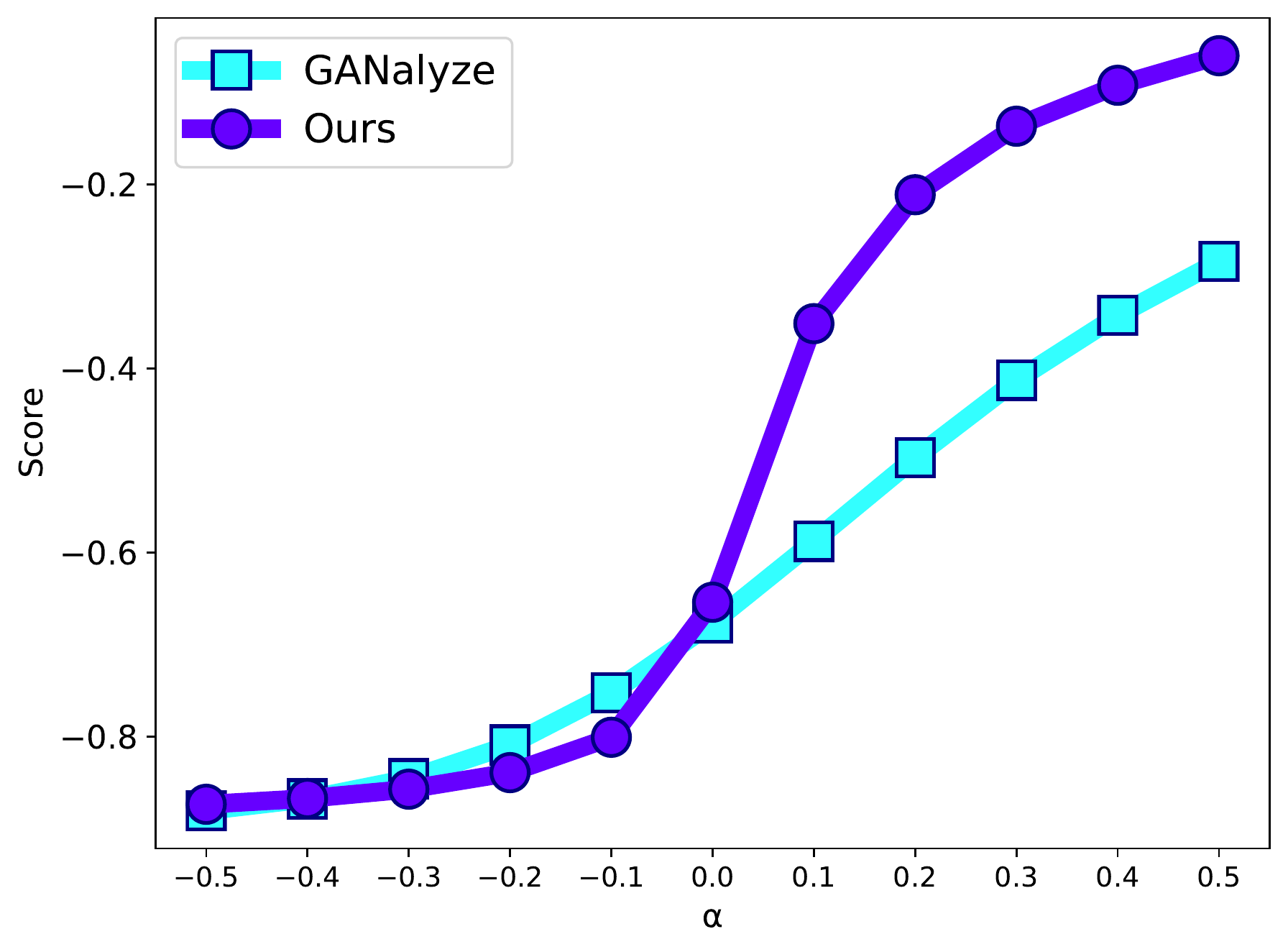} 
  \includegraphics[width=0.13\linewidth]{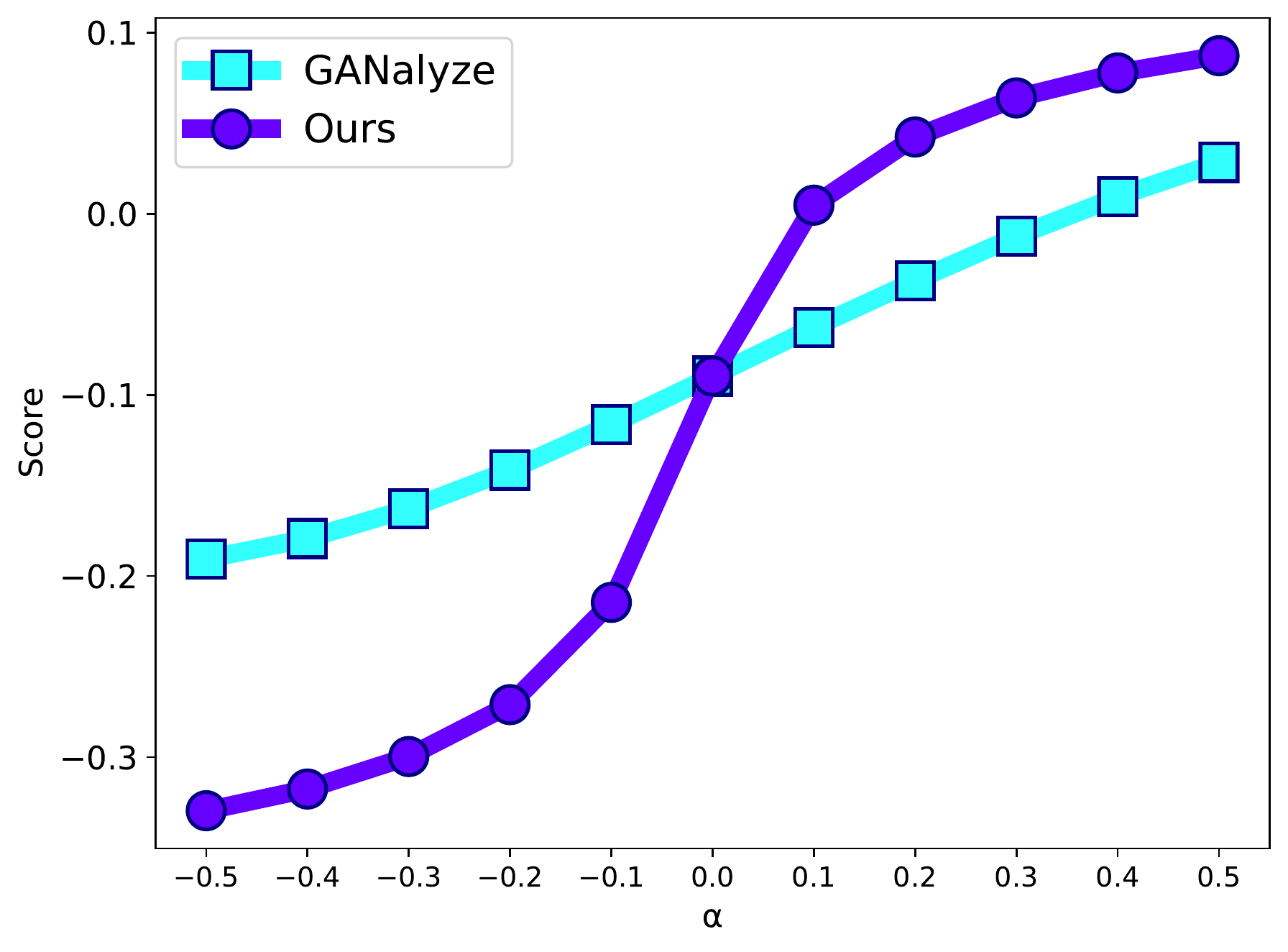} 
  \includegraphics[width=0.13\linewidth]{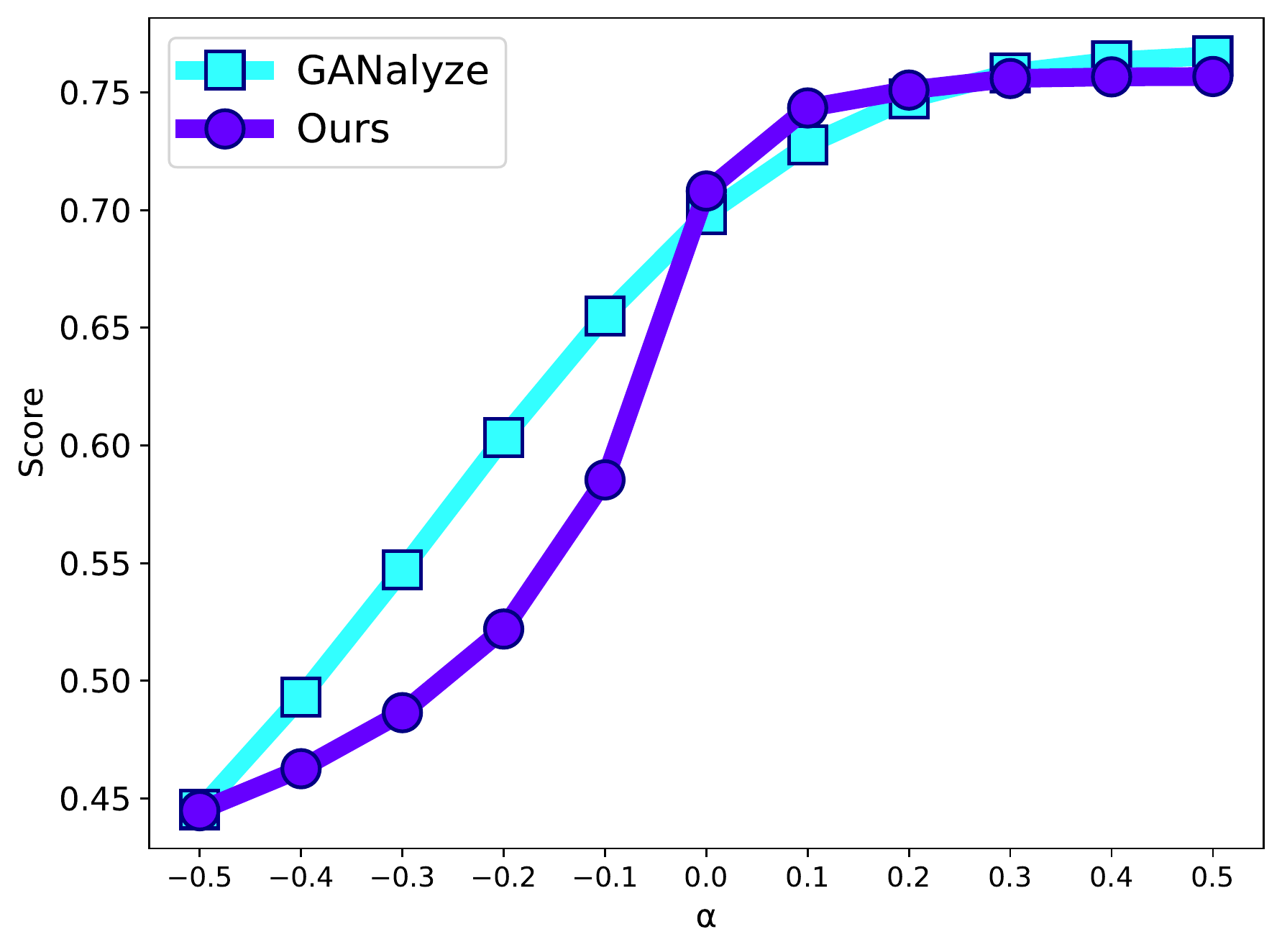} 
  \includegraphics[width=0.13\linewidth]{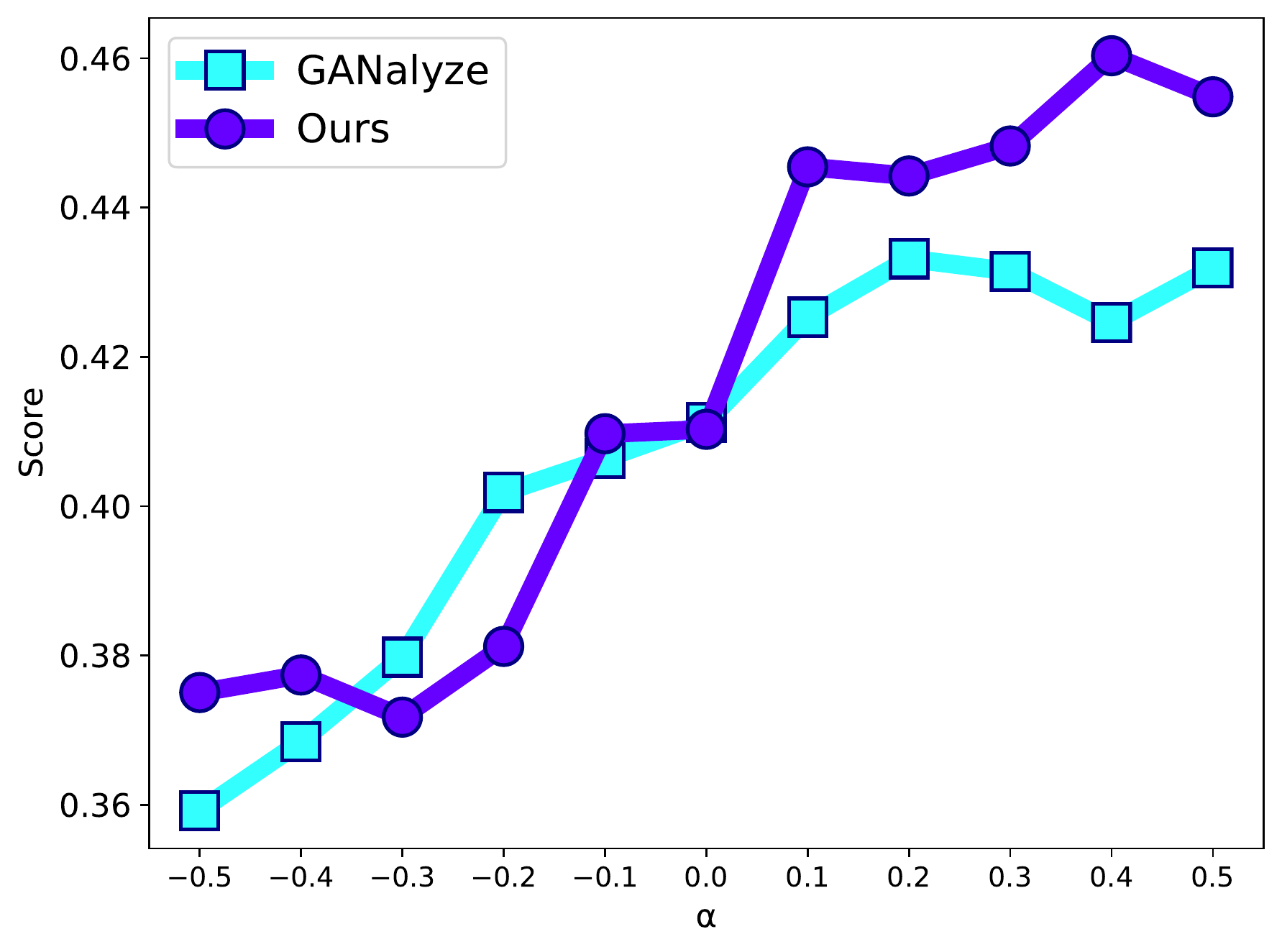} 
  \includegraphics[width=0.13\linewidth]{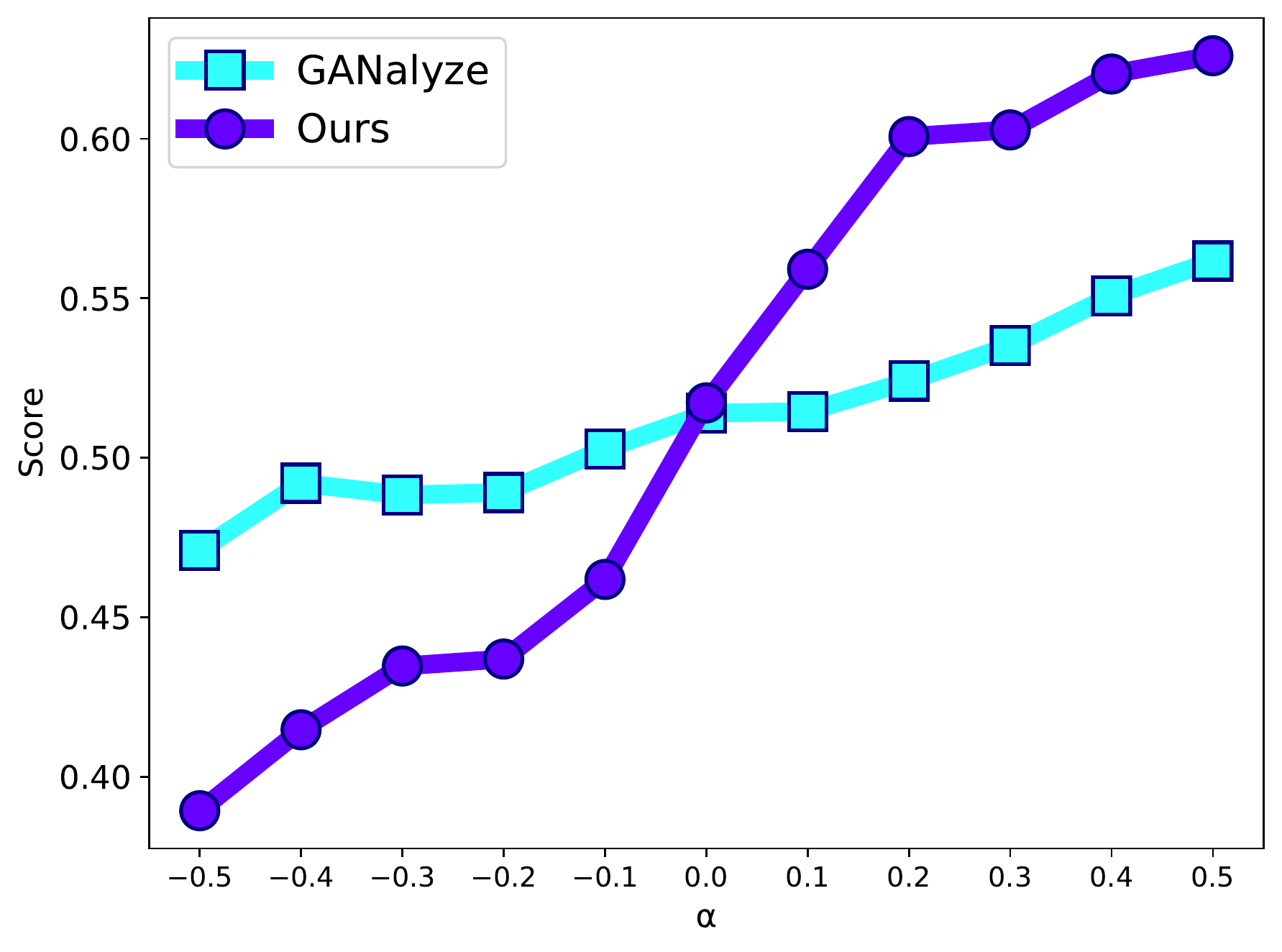} 
  \includegraphics[width=0.13\linewidth]{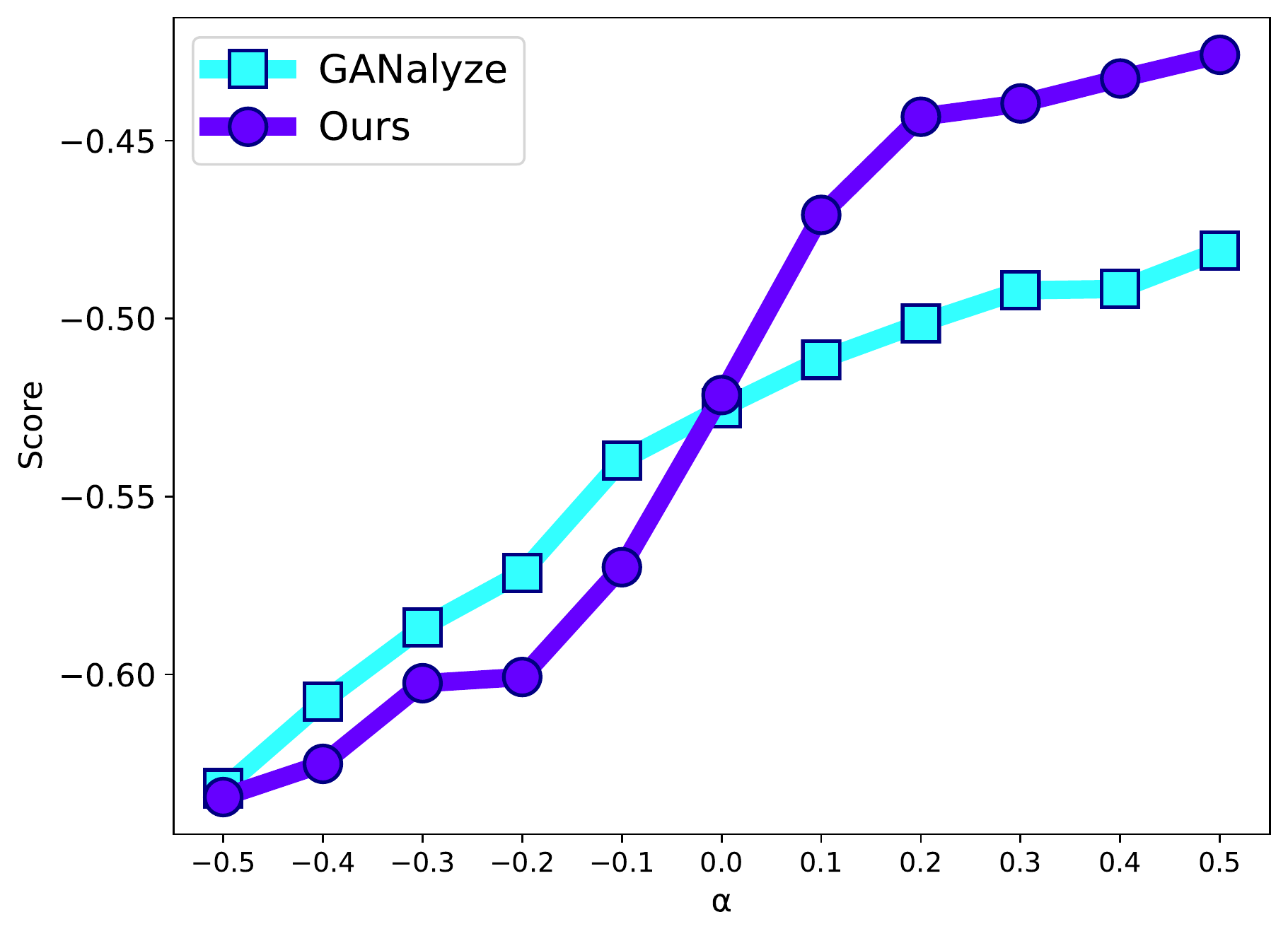}

\caption{We investigate the image factors altered using our method and GANalyze \cite{goetschalckx2019ganalyze}. We compute Redness, Colorfulness, Brightness, Entropy, Squareness, Centeredness, and Object Size metrics for Memorability (Memnet), Emotional Valence (Emonet), and Aesthetics. $x$ axis represents $\alpha$ value while $y$ axis represents the mean score of the property of interest.}
\label{fig:emerging_factors}
\end{center}
\vskip -0.2in
\end{figure*}

All methods are trained for $200K$ iterations. For experiments with multiple directions, we use $\lambda = 1/100$.  Following \cite{goetschalckx2019ganalyze}, we generate a test set of 750 images with randomly selected ImageNet classes with two $\mathbf{z}$-vectors per class. Each $\mathbf{z}$ vector is paired with 11 different $\alpha$ values $[-0.5,\ldots,-0.1,0,0.1,\ldots,0.5]$, where $\alpha = 0$ recovers the original image $\mathcal{G}(\mathbf{z},y)$. Thus, our final test set consists of $1500 \times 11$ images, for a total of $16,500$ images.

\subsubsection {Cognitive Models}
Similar to \cite{goetschalckx2019ganalyze}, our objective function requires a scoring function that assesses the extent to which a particular cognitive property such as \textit{memorability} is present in a given image. In this work, we use pre-trained scoring functions for memorability, emotional valence, and aesthetics. In the following, we describe the details of each model.

\begin{itemize} \item \textbf{Memorability (MemNet)}: Previous research \cite{khosla2015understanding} suggests that memorable images have different intrinsic visual features. Therefore, memorability is an important cognitive property that measures how memorable a particular image is. We use a pre-trained model of \cite{khosla2015understanding} that predicts the memorability of images. 

\item \textbf{Emotional Valence (EmoNet)}: Emotional valence refers to the degree  which an image evokes positive or negative emotions. We use a pre-trained Emotional Valence model of \cite{goetschalckx2019ganalyze} trained by fine-tuning a ResNet50 model \cite{he2016deep} on the Moments database \cite{monfort2019moments} and the Cornell Emotion6 database \cite{peng2015mixed}. 

\item \textbf{Aesthetics (AestheticsNet)}: \cite{kong2016photo} shows that image features are associated with positive aesthetic judgments in most individuals, and that it is possible to evaluate the aesthetics of a given image. We use a pre-trained model of \cite{goetschalckx2019ganalyze}, which uses a CNN-based model of \cite{kong2016photo}.

\end{itemize}

\subsection {Performance on Single Directions}
First, we investigate whether our model learns to navigate the latent space in such a way that it can manipulate the cognitive property of interest, i.e. \textit{memorability}, for a given $\alpha$ value using a \textit{single} direction. Figure \ref{fig:exp_cond} shows a comparison of the classifier scores based on different $\alpha$ scores. We can see that our method performs better manipulation than \cite{goetschalckx2019ganalyze} for all the cognitive features we are interested in. For example, for Emotional Valence, we see that our model achieves much lower Emotional Valence score for low $\alpha$ values such as $\alpha=-0.5$ or $\alpha=-0.4$, while it achieves higher Emotional Valence score for $\alpha=0.4$ and $\alpha=0.5$. This observation also holds for other cognitive properties, with our model scoring lower for decreasing $\alpha$ values and higher for increasing $\alpha$ values compared to  \cite{goetschalckx2019ganalyze}. 

\paragraph{Qualitative Results} A visual comparison between our method and GANalyze can be seen in Figure \ref{fig:single_outputs}. We can see that our method performs successful manipulation towards $\alpha$- - and $\alpha$++ while achieving better or similar scores to GANalyze.

 \subsection {Performance on Multiple Directions}
Next, we examine whether multiple directions achieve similar performance in terms of manipulating the cognitive properties while yielding diverse outputs. Figure \ref{fig:exp_cond} shows a comparison of the classifier scores based on different $\alpha$ scores using $k=3$ different directions generated by our model and GANalyze. We can see that all three directions obtain similar scores with increasing and decreasing $\alpha$ values. 

\paragraph{Qualitative Results} Figure \ref{fig:teaser} and Figure \ref{fig:single_outputs_multi} show the qualitative results for $k=3$ directions. For  Memorability and Emotional Valence, our framework manipulates the input images by applying different colors for the background and positioning, while maintaining a similar increase in score for a given $\alpha$ value. Similarly, for Aesthetics, our model changes the object size, background, and color scheme to manipulate the image in diverse ways while maintaining a similar increase in the scoring function.

\subsection{Statistical Analysis}
We also formally test these observations by fitting a linear mixed-effects regression model to the data. While \cite{goetschalckx2019ganalyze} achieves an unstandardized slope ($\beta$) of $0.522$ ($95\% \text{CI}= [0.44, 0.604]$, $p < 0.001$), our method achieves a slope of $0.901$ ($95\% \text{CI}= [0.704, 1.098]$, $p < 0.001$) on memorability. Our method also achieves a similar improvement in Emotional Valence. While \cite{goetschalckx2019ganalyze} has a slope of $ 0.28$ ($95\% \text{CI}=[0.272, 0.288] $, $p < 0.001$), our method has a slope of $1.017$ ($95\% \text{CI}= [0.979, 1.056] $, $p < 0.001$). Similarly, for Aesthetics, \cite{goetschalckx2019ganalyze} obtains a slope of $0.268$ ($95\% \text{CI}= [0.258, 0.278] $, $p < 0.001$), while our method has $0.615$ ($95\% \text{CI}= [0.541, 0.69] $, $p < 0.001$). We also note that the standard deviations between the scores from GANalyze and our model are very similar. For the Memorability, Emotional Valence, and Aesthetics properties, GANalyze has a standard deviation of $0.116$, $0.124$, and $0.114$, while our method achieves $0.114$, $0.111$, and $0.116$ respectively.

\begin{figure*}[ht]
\vskip 0.2in
\qquad \qquad \quad    Direction 1-2  \qquad  \qquad \qquad \qquad \qquad \quad \, Direction 1-3 \qquad \qquad \qquad \qquad \qquad \: Direction 2-3 
\vspace{-0.3cm} 
\begin{center}
\begin{turn}{90} \qquad \qquad \quad  \, Score \end{turn}
\includegraphics[width=0.65\columnwidth]{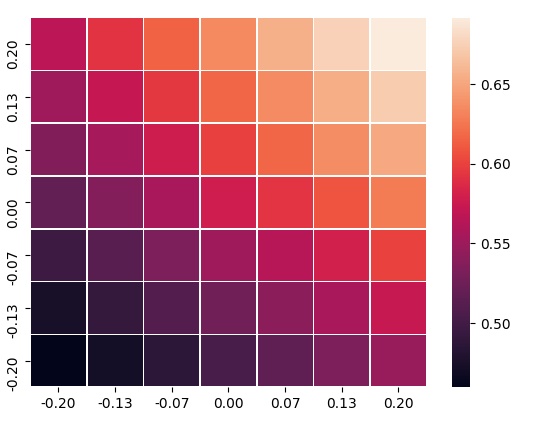}
\begin{turn}{90} \qquad \qquad \quad  \, Score \end{turn}
\includegraphics[width=0.65\columnwidth]{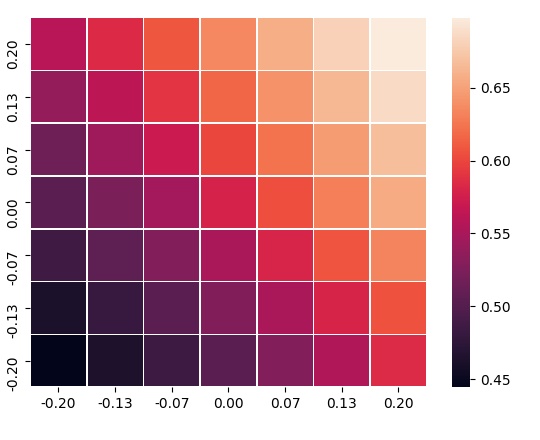}
\begin{turn}{90} \qquad \qquad \quad  \, Score \end{turn}
\includegraphics[width=0.65\columnwidth]{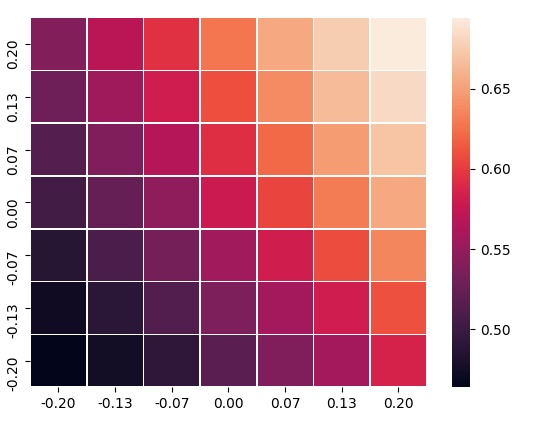}

\vspace{-0.2cm} 
\hspace{-0.4cm} Score \hspace{4.9cm} Score \hspace{4.9cm} Score

    \caption{Consistency analysis for $k=3$ diverse directions found using our method where we interpolate each combination of direction pairs (Directions \{1-2, 1-3, 2-3\}) and visualize the scores as heatmaps. We can see that all three interpolations yield consistent changes in the score.} 
\label{fig:sub1}
\end{center}
\vskip -0.2in
\end{figure*}

\subsection {Exploring Image Factors}
So far, we have shown that our method improves the manipulation ability and  produces diverse outputs. Next, we explore what kind of image factors are changed to achieve these improvements. Following \cite{goetschalckx2019ganalyze}, we explore  \textit{Redness, Colorfulness, Brightness, Entropy, Squareness, Centeredness}, and \textit{Object Size} factors, which are described as follows.

\begin{itemize}
    \item \textbf{Redness} is calculated as the normalized number of red pixels. 
    \item \textbf{Colorfulness} metric is calculated using \cite{hasler2003measuring} which considers the distribution of the image pixels in the CIELab colour space.
    \item \textbf{Brightness} is calculated as the average pixel value after the grayscale transformation.
    \item \textbf{Entropy} is calculated using the pixel intensity histogram following \cite{goetschalckx2019ganalyze}.
    \item \textbf{Object size} We used Mask R-CNN \cite{he2017mask} to create a segmentation mask at the instance level of the generated image. Then, following \cite{goetschalckx2019ganalyze} we used the difference in mask area as the $\alpha$ value increases or decreases.
    \item \textbf{Centeredness} We used the difference  between the centroid of the mask from the center of the frame.
    \item   \textbf{Squareness} We used the ratio of the length of the minor and major axes of an ellipse with the same normalized central moments as the mask.
\end{itemize}

Figure \ref{fig:emerging_factors} shows the comparison between GANalyze \cite{goetschalckx2019ganalyze} and our method on different factors. We observe similar behavior for both \cite{goetschalckx2019ganalyze} and our model where they increase \textit{Redness, Colorfulness} and \textit{Brightness} for Emotional Valence, Memorability and Aesthetics. Moreover, while our model prefers decreasing \textit{Entropy}, \cite{goetschalckx2019ganalyze} prefers increasing  \textit{Entropy} for Emotional Valence and  Memorability as $\alpha$ increases. On the other hand, our method for \textit{Emotional Valence} prefers to increase \textit{Squareness, Centeredness} and \textit{Object Size} as $\alpha$ increases. In contrast, GANalyze favors decreasing \textit{Squareness, Centeredness} and \textit{Object Size} as $\alpha$ increases. For \textit{Memorability} and \textit{Aesthetics}, both methods consistently increase \textit{Squareness, Centeredness} and \textit{Object Size} as $\alpha$ increases.

\subsection{Diversity and Consistency Analysis}
We examine how assessor scores change as a function of number of directions. Table \ref{tab} shows the cosine similarity statistics and assessor scores with $k=\{3,5,10\}$ directions.  We find that the scores increase steadily up to $k=5$ and decrease as the number of directions increases (e.g., $k=10$). As expected, the cosine distance decreases slightly as we increase the number of directions. The increase in assessor score is calculated as the average value of $k$ directions and the cosine distance values are calculated as the average distance between directions. 

We also perform a consistency analysis for $k=3$ diverse directions using our method, where we interpolate each combination of direction pairs and visualize the scores as heatmaps. $x$ and $y$ axis scores represent the $\alpha$ value of the corresponding direction. $y$ axis is the score change in the first direction and the $x$ axis is the change in the second direction. The color gives the average assessor score of the images. Our analysis in Figure \ref{fig:sub1} shows that all three interpolations yield consistent changes in scores.

\subsection{Image Quality Analysis}
We also examined the image quality using FID \cite{heusel2017gans} metric. We obtain an average FID of $59.78$, while the GANalyze method obtains $56.04$ for Emotional Valence. For Aesthetics, we obtain $56.44$ FID, while GANalyze obtains $55.25$. For Memorability, we obtain $61.56$, while GANalyze obtains $54.7$. Thus, we see that both methods obtain similar FID results while manipulating the images, but our method is slightly higher than GANalyze. We note that our method is able to manipulate the values of the scoring functions more accurately (see Figure \ref{fig:exp_cond}), and therefore makes larger changes in the images (and thus larger changes in the FID score).

\begin{table}[t!]
\begin{center}
\resizebox{0.7\columnwidth}{!}{
\begin{tabular}{|c|c|c|}
\hline
\textbf{Number of} & \textbf{Increase in} & \textbf{Cosine} \\ \textbf{Directions} & \textbf{Assessor Score} & \textbf{Distance}  \\
\hline
$k=3$ & 0.08 & 1.3   \\ \hline
$k=5$ & 0.08 & 1.2  \\\hline
$k=10$ & 0.02 & 1.1  \\
\hline
\end{tabular}
} 
\end{center}
\vspace{-0.2cm}
\caption{Changes in cosine distance and assessor scores as a function of the number of directions. }

\label{tab}
\end{table}
 \section{Limitations and Social Impact}
\label{sec:limitations}
On a technical level, our method requires an assessor function to learn directions, which may not always be available. While our method is currently limited to manipulating images generated by GAN models,  inversion methods \cite{gu2020image} can be used to manipulate real images. On an ethical level, our method shares similar concerns and dangers as those raised by \cite{korshunov2018deepfakes}. In particular, our method can be used to manipulate facial images like FFHQ \cite{StyleGAN} and can be used by malicious parties to manipulate and misuse images of real people. 

Furthermore, we acknowledge that assessing cognitive properties such as \textit{Memorability}, \textit{Emotional Valence} and \textit{Aesthetics} is an extremely complex task, and the pre-trained assessor functions \cite{khosla2015understanding, kong2016photo, monfort2019moments, peng2015mixed} used in this study may carry some biases.

 \section{Conclusion}
\label{sec:conclusion}
In this study, we propose a framework that learns multiple and diverse manipulation for cognitive properties. We applied it to \textit{Memorability, Emotional Valence, Aesthetics} and show that our method performs a better manipulation than \cite{goetschalckx2019ganalyze} while producing diverse outputs. We  conduct qualitative and quantitative experiments to investigate the performance of our method in terms of single and multiple direction capability, statistical analysis, diversity, and image quality studies. We also investigate the visual features such as \textit{redness, colorfulness, object size} that contribute to achieving the desired manipulations.  

Our source code can be found at \url{http://catlab-team.github.io/latentcognitive}.\\

\noindent \textbf{Acknowledgments}
This publication has been produced benefiting from the 2232 International Fellowship for Outstanding Researchers Program of TUBITAK (Project No: 118c321). 

{\small
\bibliographystyle{ieee_fullname}
\bibliography{example_paper}
}

\end{document}